\documentclass[letterpaper, 10 pt, conference]{ieeeconf}  
\IEEEoverridecommandlockouts                              

\title{\bf Decision making in dynamic and interactive environments based on cognitive hierarchy theory, Bayesian inference, and predictive control}

\author{Sisi Li, Nan Li, Anouck Girard, and Ilya Kolmanovsky 
    \thanks{This research has been supported by the National Science Foundation award CNS 1544844.}
    \thanks{
        Sisi Li, Nan Li, Anouck Girard, and Ilya Kolmanovsky are with the Department of Aerospace Engineering,
        University of Michigan, Ann Arbor, MI 48109, USA
        {\tt\small \{sisli, nanli, anouck, ilya\}@umich.edu}.}
}
\usepackage{float}
\usepackage[caption=false]{subfig}
\usepackage{epsfig}
\usepackage{epstopdf}
\usepackage{bbm}
\usepackage{epsfig} 
\usepackage{amsmath} 
\usepackage{amssymb}  
\usepackage{bm}
\usepackage{epstopdf}
\usepackage{color}
\usepackage{algorithm}
\usepackage[noend]{algpseudocode}
\DeclareMathOperator*{\argmax}{arg\,max}

\usepackage{comment}
\begin{document}
\maketitle

\begin{abstract}
In this paper, we describe an integrated framework for autonomous decision making in a dynamic and interactive environment. We model the interactions between the ego agent and its operating environment as a two-player dynamic game, and integrate cognitive behavioral models, Bayesian inference, and receding-horizon optimal control to define a dynamically-evolving decision strategy for the ego agent. Simulation examples representing autonomous vehicle control in three traffic scenarios where the autonomous ego vehicle interacts with a human-driven vehicle are reported.
\end{abstract}



%
\IEEEpeerreviewmaketitle

\section{Introduction}\label{sec:Intro}

Many autonomous systems, including collaborative robots \cite{guiochet2017safety} and self-driving cars \cite{schwarting2018planning}, operate in dynamic and interactive environments. As an example, a self-driving car may operate in traffic with multiple other cars, which inform the environment. The environment is dynamic and interactive because the other cars not only operate concurrently with the ego car but also respond to its actions \cite{li2018game}. It is important for such autonomous systems to account for these interactions in decision making to ensure safe and effective operations, especially for systems interacting with humans. Despite much progress made, this problem remains challenging and unsolved in many application scenarios \cite{schwarting2018planning,goodrich2008human}.

Strategic interactions between intelligent agents can be modeled using game theoretic frameworks \cite{myerson2013game}, among which cognitive hierarchy theory has drawn attention from game theorists and practitioners since the 90's \cite{nagel1995unraveling,stahl1995players} due to its potential for more accurate prediction of human behavior compared to equilibrium-based theories shown in many experimental studies \cite{costa2006cognition,costa2009comparing}. Cognitive hierarchy theory describes human thought processes in strategic games by characterizing human behavior based on levels of iterated rationalizability. In particular, it assumes bounded rationality of decision makers in contrast to the assumption of unbounded/perfect rationality in many equilibrium-based theories. The assumption of bounded rationality can be more realistic than that of unbounded rationality in many practical situations because the reasoning capability of a decision maker is often limited by the complexity of the decision problem and the time available to make a decision \cite{gigerenzer2002bounded}.

In this paper, we describe a decision making framework for autonomous systems, which integrates cognitive hierarchy theory, Bayesian inference of cognitive level, and receding-horizon optimization applied to a partially observable Markov decision process formulation to define a dynamically-evolving decision strategy, where hard constraints can also be imposed and probabilistically enforced over the planning horizon to incorporate safety requirements.

We note that although cognitive hierarchy theory has been utilized for modeling multi-agent interactions in the literature \cite{li2018game,yildiz2014predicting,musavi2016unmanned,kanellopoulos2019non}, most of the existing works exploit a ``level-k thinking'' framework \cite{nagel1995unraveling,stahl1995players}, which assumes that a level-$k$ decision maker treats the interactive environment as a level-($k-1$) decision maker and responds to it accordingly. Unfortunately, level-$k$ decision rules may lead to poor decisions if the ($k-1$)-assumption about the interactive environment's cognitive level is incorrect \cite{li2018game_intersection}.

Therefore, we consider another framework that is related to ``level-k thinking'' but has distinguishing features, called the ``cognitive hierarchy'' framework \cite{camerer2004cognitive}, where CH-$k$ decisions are optimized to strategically respond to the interactive environment by modeling it using a mixture of level-$\sigma$, $\sigma = 0,\cdots,k-1$, decision-maker models. The cognitive hierarchy framework enables the autonomous agent to strategically interact with environments with different cognitive levels. When the autonomous agent has good knowledge about its operating environment, the mixing ratio of level-$\sigma$ models could be pre-specified \cite{abuzainab2016cognitive}. When operating in an uncertain environment, reasoning about the interactive environment's cognitive level could be incorporated in the decision making process.
In particular, unlike the approaches in \cite{kanellopoulos2019non,li2018game_intersection,tian2018adaptive} relying on heuristic techniques to obtain level estimates, the level reasoning algorithm in this paper is based on Bayesian inference, which may have a broader applicability.

This paper is organized as follows: In Section~\ref{sec:PF}, we formulate the decision making problem of an autonomous agent operating in a dynamic and interactive environment as a dynamic game. In Section~\ref{sec:CHT}, we review two related but distinct frameworks of cognitive hierarchy theory, the level-k framework and the cognitive hierarchy framework. In Section~\ref{sec:POMDP}, we describe our integration of cognitive hierarchy theory, Bayesian inference of cognitive level, and receding-horizon optimization applied to a constrained partially observable Markov decision process formulation to determine the ego agent's decision strategy. In Section~\ref{sec:AV}, we apply the proposed decision making framework to autonomous vehicle control in three interactive traffic scenarios. Discussion and conclusions are given in Section~\ref{sec:conclusion}.

\section{Problem formulation}\label{sec:PF}

In this paper, we consider a decision making process by an intelligent agent operating in a dynamic and interactive environment. The interactions between the ego agent and the environment are modeled as a two-player dynamic game represented as a 6-tuple,
\begin{equation}
\big\langle P, X, U, T, R, C \big\rangle,
\end{equation}
where $P = \{1,2\}$ represent the two players with $1$ denoting the ego agent and $2$ denoting the environment; $X$ is a finite set of states with $x_t \in X$ denoting the state of the agent-environment system at the discrete time instant $t$; $U = U^1 \times U^2$ is a finite set of actions with $U^1$ denoting the action set of the ego agent and $U^2$ denoting the action set of the environment; $T$ represents a transition of the state $x_t \to x_{t+1}$ as a result of an action pair $(u^1_t,u^2_t) \in U$, in particular, $T$ is defined by the following dynamic model,
\begin{equation}\label{equ:dynamic_1}
x_{t+1} = T(x_t,u^1_t,u^2_t);
\end{equation}
$R = \{R^1,R^2\}$ are reward functions of the two players representing their decision objectives, in particular,
\begin{equation}
R^i_t = R^i(x_t,u^1_t,u^2_t), \quad i \in P,
\end{equation}
i.e., each player's reward at the time instant $t$ depends on the state $x_t$ and both players' actions $(u^1_t,u^2_t)$; and $C = \{X_t\}_{t \in \mathbb{N}}$ with $X_t \subseteq X$ being a set of ``safe'' states, representing hard constraints for decision making of the ego agent.

We consider a receding-horizon optimization-based process for the decisions of the ego agent,
\begin{subequations}\label{equ:DM_1}
\begin{align}
    &\, ({\text{\bf u}}_t^1)^* = \big\{(u_{0|t}^1)^*,\cdots,(u_{N-1|t}^1)^*\big\} \in \label{equ:action_1} \\[-2pt]
    &\, \argmax_{u_{\tau|t}^1 \in U^1,\, \tau = 0,\cdots, N-1}\, \sum_{\tau = 0}^{N-1} \lambda^\tau R^1 \big(x_{\tau|t}, u_{\tau|t}^1, u_{\tau|t}^2\big), \nonumber \\
    &\text{ s.t.}\quad x_{\tau+1|t} \in X_{t+\tau+1}, \quad \forall\, \tau = 0, \cdots, N-1, \label{equ:constraint_1}
\end{align}
\end{subequations}
where $u_{\tau|t}^i$ denotes a prediction of player $i$'s action at the time instant $t+\tau$ over a planning horizon of length $N$ with the prediction made at the current time instant $t$, $x_{\tau|t}$ denotes a prediction of the system state under the sequence of action pairs $\big\{(u_{\tau|t}^1,u_{\tau|t}^2)\big\}_{\tau = 0}^{N-1}$, and $\lambda \in (0,1]$ is a factor to discount future rewards.

Clearly, the above optimization problem is not well-defined yet, as the uncontrolled variables $\big\{u_{0|t}^2,\cdots,u_{N-1|t}^2\big\}$ are unknown. One way to proceed is to consider worst-case scenarios, i.e., replace \eqref{equ:action_1} with
\begin{equation}\label{equ:DM_maximin}
    ({\text{\bf u}}_t^1)^* \in \argmax_{u_{\tau|t}^1 \in U^1} \min_{u_{\tau|t}^2 \in U^2} \,\, \sum_{\tau = 0}^{N-1} \lambda^\tau R^1 \big(x_{\tau|t}, u_{\tau|t}^1, u_{\tau|t}^2\big).
\end{equation}
However, as \eqref{equ:DM_maximin} assumes an adversarial player~$2$, rather than a rational player~$2$ that pursues its own objectives and is not necessarily against the ego agent, \eqref{equ:DM_maximin} may lead to overly-conservative decisions for the ego agent. Therefore, we pursue an alternative solution, which is to exploit cognitive hierarchy theory to predict player $2$'s actions.

\section{Two cognitive hierarchy frameworks}\label{sec:CHT}

Cognitive hierarchy theory is concerned with behavioral models describing human thought processes in strategic games. It characterizes human behavior based on levels of iterated rationalizability. Two frameworks have been developed based on cognitive hierarchy theory: the level-k framework \cite{nagel1995unraveling,stahl1995players} and the cognitive hierarchy framework \cite{camerer2004cognitive}, which are related but have distinct features. They are reviewed in what follows.

\subsection{The level-k framework}
\label{sec: level-k}

In the level-k framework, it is assumed that each player in a strategic game bases its decision on a finite depth of reasoning about the likely actions of the other players, which is referred to as its ``cognitive level.'' In particular, the reasoning hierarchy starts from some non-strategic behavioral model, called level-$0$. Then, a level-$k$ player, $k=1,2,\cdots$, assumes that all other players are level-($k-1$), based on which it predicts the actions of the other players and makes its own decision as the optimal response to these predicted actions. In short, a level-$k$ decision optimally responds to level-($k-1$) decisions. 

The level-k framework has been exploited for modeling human-human, human-machine, and machine-machine\footnote{On the basis of the fact that many machine systems pursue human-like decision making, e.g., in \cite{yu2018human}.} interactions in automotive systems \cite{li2018game, li2018game_intersection,tian2018adaptive}, aerospace systems \cite{yildiz2014predicting,musavi2016unmanned}, as well as in cyber-physical security \cite{kanellopoulos2019non}. It has been revealed in \cite{li2018game_intersection} that an ego agent using a decision strategy corresponding to a level-$k$ model with some fixed $k$ may behave poorly when interacting with another agent using a decision strategy of level-$\sigma$ with $\sigma \neq k-1$. 

\subsection{The cognitive hierarchy framework}
\label{sec: CH framework}

The cognitive hierarchy (CH) framework is similar to the level-k framework in terms of characterizing each player's behavior also by a cognitive level $k$. The unique feature of the CH framework is the hypothesis that a player can act under the assumption that some percentage of the population fits each archetype. More specifically, a CH-$k$ player assumes that each of the other players is level-$\sigma$ for some $\sigma < k$ and optimizes its decision corresponding to its beliefs about the other players' levels. This feature makes a CH-$k$ player ``smarter'' than a level-$k$ player by enabling a CH-$k$ player to optimally respond to level-$\sigma$ decisions for all $\sigma < k$ as long as it has the correct beliefs about the other players' levels.

On the basis of this observation, algorithms that estimate the other agents' cognitive levels according to their historical behavior have been proposed in \cite{kanellopoulos2019non,li2018game_intersection,tian2018adaptive} so that the ego agent can adapt its decision strategy to the level estimates.


\section{Decision making framework exploiting cognitive hierarchy theory}\label{sec:POMDP}

In this section, we describe our decision making framework that integrates cognitive hierarchy theory, Bayesian inference, and receding-horizon optimal control.

\subsection{Level-k models of the environment}

A policy $\pi^i$, $i \in P$, is a stochastic map from states $X$ to actions $U^i$. Specifically, $\pi^i: X \times U^i \to [0,1]$ is such that
\begin{equation}
\mathbb{P}(u^i_t = u|x_t = x) = \pi^i(x,u),
\end{equation}
for all $t \in \mathbb{N}$, where $\mathbb{P}(\cdot|\cdot)$ denotes conditional probability.

To define the level-$k$ models of the environment for arbitrary $k = 0,1,\cdots$, we start from defining a level-$0$ model of the ego agent, defined by a policy $\pi^{1,0}$, and a level-$0$ model of the environment, defined by a policy $\pi^{2,0}$. The level-$k$ model of the environment, $\pi^{2,k}$, with $k \ge 1$ is then constructed based on the ``softmax decision rule'' \cite{sutton2018reinforcement}, which captures the suboptimality and variability in decision making \cite{beck2012not,acerbi2014origins}, as follows,
\begin{equation}\label{equ:levelk_policy}
\pi^{2,k}(x,u) = \frac{\exp\big(Q^{2,k}(x,u)\big)}{\sum_{u' \in U^2} \exp\big(Q^{2,k}(x,u')\big)},
\end{equation}
in which the $Q$-function of state-action pairs is defined as
\begin{align}
& Q^{2,k}(x,u) = \max_{u_{\tau|t}^2 \in U^2,\, \tau = 1,\cdots, N-1} \\[-6pt]
& \mathbb{E}\Big\{ \sum_{\tau = 0}^{N-1} \lambda^\tau R^2 \big(x_{\tau|t}, u_{\tau|t}^1, u_{\tau|t}^2\big)\,\Big|\, x_{0|t} = x,u_{0|t}^2=u, \pi^{1,k-1}\Big\}, \nonumber
\end{align}
where $\pi^{1,k-1}$ is the level-($k-1$) model of the ego agent, which for $k \ge 2$ is defined as
\begin{equation}
\pi^{1,k-1}(x,u) = \frac{\exp\big(Q^{1,k-1}(x,u)\big)}{\sum_{u' \in U^1} \exp\big(Q^{1,k-1}(x,u')\big)},
\end{equation}
in which
\begin{align}
& Q^{1,k-1}(x,u) = \max_{u_{\tau|t}^1 \in U^1,\, \tau = 1,\cdots, N-1} \\[-6pt]
& \mathbb{E}\Big\{ \sum_{\tau = 0}^{N-1} \lambda^\tau R^1 \big(x_{\tau|t}, u_{\tau|t}^1, u_{\tau|t}^2\big)\,\Big|\, x_{0|t} = x,u_{0|t}^1=u, \pi^{2,k-2}\Big\}. \nonumber
\end{align}

In short, the level-$k$ model of the environment is constructed based on the level-($k-1$) model of the ego agent, which is constructed based on the level-($k-2$)
model of the environment. Therefore, the level-$k$ models of the environment as well as the level-$k$ models of the ego agent are constructed recursively for $k=0,1,\cdots$. We note that when constructing the level-$k$ models of the ego agent, we drop the hard constraints \eqref{equ:constraint_1} to reduce computational complexity but can promote their satisfaction through imposing penalties in the reward function $R^1$.

\subsection{Decision making based on the CH framework}

After the level-$k$ models of the environment $\pi^{2,k}$ for $k = 0,1,\cdots,k_{\max}$ have been constructed, we define the augmented state of the agent-environment system as $\overline{x}_t = [x_t,\, \sigma]^\top$, where $\sigma \in K = \{0,1,\cdots,k_{\max}\}$ represents the actual cognitive level of the environment and is assumed to be unknown to the ego agent. Then, we consider the following augmented dynamic model of the agent-environment system,
\begin{subequations}
\begin{align}
    \overline{x}_{t+1} &=  f(\overline{x}_t,u^1_t,u^2_t) = \begin{bmatrix} T(x_t,u^1_t,u^2_t) \\ \sigma \end{bmatrix}, \label{equ:dynamic_2} \\
    y_t &= g(\overline{x}_t) = x_t, \label{equ:dynamic_21}
\end{align}
\end{subequations}
where $y_t = x_t \in X$ is referred to as an ``observation.'' On the basis of the level-$k$ policies, the action of the environment, $u_t^2$ in \eqref{equ:dynamic_2}, can be viewed as a stochastic disturbance satisfying
\begin{equation}\label{equ:disturbance}
\mathbb{P}(u_t^2 = u|x_t = x, \sigma = k) = \pi^{2,k}(x,u),
\end{equation}
for all $u \in U^2$ and $(x,k) \in X \times K$.

It is assumed that the ego agent has a prior belief about $\sigma$, as a probability distribution, $\mathbb{P}(\sigma = k)$, defined over $K = \{0,1,\cdots,k_{\max}\}$. Let us collect all historical observations up to time $t$ and all previously executed actions by the ego agent up to time $t-1$ into a data vector,
\begin{equation}
\xi_t = \big\{y_0,\cdots,y_t,u_0^1,\cdots,u_{t-1}^1 \big\},
\end{equation}
which, roughly speaking, will be used by the ego agent as evidence to infer the actual cognitive level of the environment, i.e., to obtain the posterior belief about $\sigma$, $\mathbb{P}(\sigma = k | \xi_t)$, defined for all $k \in K$.

Then, we consider the following decision making process by the ego agent,
\begin{subequations}\label{equ:DM_2}
\begin{align}
    &\!\!\!\! ({\text{\bf u}}_t^1)^* \!\in \argmax_{u_{\tau|t}^1 \in U^1}\, \mathbb{E} \Big\{\! \sum_{\tau = 0}^{N-1}\! \lambda^\tau R^1 \big(x_{\tau|t}, u_{\tau|t}^1, u_{\tau|t}^2\big) \,\Big|\, \xi_t \Big\}, \! \label{equ:action_2} \\
    &\!\!\!\!\!\! \text{ s.t.}  \,\, \mathbb{P} \big\{x_{\tau+1|t} \in X_{t+\tau+1},\, \forall\, \tau = 0, \cdots, N-1 \,\big|\, \xi_t \big\} \ge 1 - \varepsilon, \label{equ:constraint_2}
\end{align}
\end{subequations}
where $\varepsilon \in [0,1]$ defines a required level of confidence in constraint satisfaction.

Comparing the processes \eqref{equ:DM_1} and \eqref{equ:DM_2}, we can observe two major differences: Firstly, the unknowns $\big\{u_{0|t}^2,\cdots,u_{N-1|t}^2\big\}$ in \eqref{equ:DM_1} have been modeled as stochastic disturbances in \eqref{equ:DM_2}, which is achieved in \eqref{equ:disturbance} by exploiting the level-$k$ models of the environment $\pi^{2,k}$. Secondly, to account for the stochasticities, the objective has been changed from maximizing the value of a function in \eqref{equ:action_1} to maximizing the expected value of the function in \eqref{equ:action_2}, and the hard constraint \eqref{equ:constraint_1} has been changed to a probabilistic requirement of satisfaction, i.e., the chance constraint \eqref{equ:constraint_2} with $\varepsilon \in [0,1]$ being a design parameter. We note that \eqref{equ:DM_2} is a well-defined optimization problem, and we present a solution method for it based on the algorithm developed in \cite{li2018tractable,li2019stochastic} in the following section.

\subsection{Solution method}

Considering randomized decision rules, we transform the optimization problem \eqref{equ:DM_2} defined in the decision space $(U^1)^N$ to the optimization problem \eqref{equ:DM_3} defined in the space of probabilities as follows:

Firstly, we define $\gamma^1_{\tau|t}$, $\tau = 0,\cdots,N-1$, as a probability distribution over the set $U^1$, based on which the predicted action $u^1_{\tau|t}$ is chosen, i.e.,
\begin{equation}
   \mathbb{P}(u^1_{\tau|t} = u_l) = \gamma^1_{\tau|t}(u_l), \quad \forall\, u_l \in U^1.
\end{equation}
Then, we reformulate \eqref{equ:DM_2} as the following optimization problem:
\begin{subequations}\label{equ:DM_3}
\begin{align}
    &\, (\Gamma_t^1)^* = \big\{(\gamma_{0|t}^1)^*,\cdots,(\gamma_{N-1|t}^1)^*\big\} \in \label{equ:action_3} \\[-2pt]
    &\, \argmax_{\gamma^1_{\tau|t} \in \Delta^1,\, \tau = 0,\cdots, N-1} \mathbb{E} \Big\{\sum_{\tau = 0}^{N-1} \lambda^\tau R^1 \big(x_{\tau|t}, u_{\tau|t}^1, u_{\tau|t}^2\big) \,\Big|\, \xi_t \Big\}, \nonumber \\
    & \text{ s.t.}  \,\, \mathbb{P} \big\{x_{\tau+1|t} \in X_{t+\tau+1},\, \forall\, \tau = 0, \cdots, N-1 \,\big|\, \xi_t \big\} \ge 1 - \varepsilon, \label{equ:constraint_3}
\end{align}
\end{subequations}
where $\Delta^1 = \big\{\gamma \in [0,1]^{|U^1|}: \sum_{i=1}^{|U^1|} \gamma_i = 1 \big\}$ is the ($|U^1|-1$)-dimensional probability simplex.

The problem \eqref{equ:DM_2} along with its transformation \eqref{equ:DM_3} is referred to as a partially observable Markov decision process (POMDP) with a time-joint chance constraint, where the partial observability comes from the unobservability of the hidden state $\sigma \in K$. A solution method for general problems in the form of \eqref{equ:DM_3} has been described in \cite{li2018tractable,li2019stochastic}. In particular, the following Propositions~1~and~2, which represent matrix-computational implementations of the corresponding mathematical expressions in \cite{li2018tractable,li2019stochastic} applied to the specific problem setting of this paper, are used in solving \eqref{equ:DM_3}.

{\bf Proposition~1:}
Suppose that the reward function $R^1$ can be written as $R^1 \big(x_{\tau|t}, u_{\tau|t}^1, u_{\tau|t}^2\big) = R^1 \big(x_{\tau+1|t} \big)$, where $x_{\tau+1|t}$ represents the next state transitioned from $\big(x_{\tau|t}, u_{\tau|t}^1, u_{\tau|t}^2 \big)$ through the dynamic model \eqref{equ:dynamic_1}. Then, for any given $\Gamma_t^1$, it holds that
\begin{equation}
\mathbb{E} \Big\{\sum_{\tau = 0}^{N-1} \lambda^\tau R^1 \big(x_{\tau+1|t} \big) \,\Big|\, \xi_t \Big\} = \big({\bf R}^1\big)^\top \Big(\sum_{\tau = 0}^{N-1}  \lambda^\tau \pi_{\tau+1|t} \Big),
\end{equation}
where ${\bf R}^1 = \begin{bmatrix}\, R^1(\overline{x}_1), \,\cdots,\, R^1(\overline{x}_{|X \times K|}) \,\end{bmatrix}^\top$ is a vector collecting the reward values associated with every $\overline{x}_i \in X \times K$, in which $R^1(\overline{x}) = R^1\big((x,k)\big) = R^1(x)$, and $\pi_{\tau+1|t} = \begin{bmatrix} \,\mathbb{P}(\overline{x}_{\tau+1|t}=\overline{x}_1 | \xi_t), \,\cdots,\,\mathbb{P}(\overline{x}_{\tau+1|t}=\overline{x}_{|X \times K|\,} | \xi_t) \,\end{bmatrix}^\top$
is a vector representing the predicted distribution of the augmented state.

{\bf Proof:} The result follows from
\begin{align*}
& \mathbb{E} \Big\{\sum_{\tau = 0}^{N-1} \lambda^\tau R^1 \big(x_{\tau+1|t} \big) \,\Big|\, \xi_t \Big\} = \sum_{\tau = 0}^{N-1} \lambda^\tau \mathbb{E} \Big\{R^1 \big(\overline{x}_{\tau+1|t} \big) \,\Big|\, \xi_t \Big\} \\
&= \sum_{\tau = 0}^{N-1} \lambda^\tau \Big(\,\sum_{\overline{x} \in X \times K}\!\!\!\! R^1 \big(\overline{x}\big)\, \mathbb{P}\big(\overline{x}_{\tau+1|t} = \overline{x} \big|\, \xi_t \big)\Big),
\end{align*}
and the definitions of ${\bf R}^1$ and $\pi_{\tau+1|t}$. $\blacksquare$

In particular, the reward vector ${\bf R}^1$ is constructed offline and the distribution vector $\pi_{\tau+1|t}$ is computed online using the recursive formula,
\begin{align}\label{equ:state_prediction}
\pi_{\tau+1|t} &= \big(I_{|X \times K|} \otimes (\gamma_{{\tau}|t})^{\top}\big) \, \text{diag}\big({\bf P}(\overline{x}_1), \,\cdots, \\
&\quad\quad\quad\quad\,\, {\bf P}(\overline{x}_{|X \times K|})\big) \, \big({\bf 1}_{|X \times K|} \otimes \pi_{\tau|t}\big), \nonumber
\end{align}
in which $I_{|X \times K|}$ denotes the $|X \times K|$-dimensional identity matrix, ${\bf 1}_{|X \times K|}$ denotes the $|X \times K|$-dimensional all-ones vector, $\otimes$ represents the Kronecker product, and
\begin{align}
& {\bf P}(\overline{x}_i) = \begin{bmatrix}\, {\bf P}(\overline{x}_i|u_1), \,\cdots,\, {\bf P}(\overline{x}_i|u_{|U^1|})\, \end{bmatrix}^\top, \\
& {\bf P}(\overline{x}_i|u_l) = \begin{bmatrix}\, \mathbb{P}(\overline{x}_i|\overline{x}_1,u_l), \,\cdots,\, \mathbb{P}(\overline{x}_i|\overline{x}_{|X \times K|},u_l)\, \end{bmatrix}^\top, \nonumber
\end{align}
with $\mathbb{P}(\overline{x}_i|\overline{x}_j,u_l)$ representing the transition kernel of the augmented state, constructed offline as
\begin{align}
&\mathbb{P}\big((x_i,k_i) \big| (x_j,k_j),u_l \big) \\
=& \sum_{u^2 \in U^2} \Big( \mathbb{I}_{\{(x_i,k_i)\}}\big(T(x_j,u_l,u^2),k_j\big)\, \pi^{2,k_j}(x_j,u^2) \Big), \nonumber
\end{align}
with $\mathbb{I}_{A}(b)$ denoting the set-membership indicator function.

The recursive formula \eqref{equ:state_prediction} starts with the initial term $\pi_{0|t}$, the posterior belief about the augmented state inferred according to the evidence $\xi_t$, which is updated at every decision step using the Bayesian inference formula,
\begin{align}\label{equ:Bayesian}
& \pi_{0|t}(\overline{x}_i) = \frac{\mathbb{I}_{\{y_t\} \times K}(\overline{x}_i)\, {\bf P}(\overline{x}_i|u^1_{t-1})^\top \pi_{0|t-1}}{\sum_{j=1}^{|X \times K|} \big(\mathbb{I}_{\{y_t\} \times K}(\overline{x}_j)\, {\bf P}(\overline{x}_j|u^1_{t-1})^\top \pi_{0|t-1} \big)}.
\end{align}

{\bf Proposition~2:} For any given $\Gamma_t^1$, the left-hand side of the constraint \eqref{equ:constraint_3} can be evaluated using the following algorithm:

\begin{enumerate}
  \item Initialize $\tau = 0$, $p^v = 0$, and $\pi^v_{0|t} = \pi_{0|t}$.
  \item Update \begin{align}
\pi^v_{\tau+1|t} &= \big(I_{|X \times K|} \otimes (\gamma_{{\tau}|t})^{\top}\big) \, \text{diag}\big({\bf P}(\overline{x}_1), \,\cdots, \nonumber \\
&\quad\quad\quad\quad\,\, {\bf P}(\overline{x}_{|X \times K|})\big) \, \big({\bf 1}_{|X \times K|} \otimes \pi^v_{\tau|t}\big). \nonumber
\end{align}
  \item Update \begin{equation} p^v \leftarrow p^v + \sum_{\overline{x} \,\notin\, X_{t+\tau+1} \times K} \pi^v_{\tau+1|t}(\overline{x}). \nonumber
  \end{equation}
  \item If $\tau = N-1$ then go to Step 5); otherwise update
  \begin{equation} \pi^v_{\tau+1|t}(\overline{x}) \leftarrow 0,\quad \forall\, \overline{x} \,\notin\, X_{t+\tau+1} \times K, \nonumber
  \end{equation}
  $\tau \leftarrow \tau+1$ and go to Step 2).
  \item Output \begin{equation}\!\! \mathbb{P} \big\{x_{\tau+1|t} \in X_{t+\tau+1},\, \forall\, \tau = 0, \cdots, N-1 \,\big|\, \xi_t \big\} = 1-p^v. \nonumber \end{equation}
\end{enumerate}

{\bf Proof:} The steps 2), 3) and 4) realize, respectively, the recursive formulas (29), (28), and (30) of Theorem~1 in \cite{li2018tractable}. Therefore, the proof follows from Theorem 1~of \cite{li2018tractable}. $\blacksquare$

On the basis of Propositions~1 and 2, a standard nonlinear programming solver exploiting gradient and Hessian information of the cost and constraint functions of \eqref{equ:DM_3}, which can be numerically estimated based on function evaluations for any $\Gamma_t^1$ of interest, can be used to solve for $(\Gamma_t^1)^*$.

\section{Application to autonomous driving in interactive traffic scenarios}\label{sec:AV}

In the near to medium term, autonomous vehicles will operate in traffic together with human-driven vehicles. Ensuring safety while maintaining performance in the associated interactive traffic scenarios remains a challenging problem for autonomous vehicle control \cite{schwarting2018planning}. In this section, we apply the proposed decision making framework to controlling an autonomous ego vehicle in traffic scenarios where it needs to interact with a human-driven vehicle. The traffic scenarios we consider include a four-way intersection scenario, a highway overtaking scenario, and a highway forced merging scenario.

Decision making of the human-driven vehicle is modeled based on the level-k framework described in Section~\ref{sec: level-k}. Experimental studies \cite{costa2006cognition,costa2009comparing} suggest that humans are most commonly level-$1$ or level-$2$ reasoners. Therefore, we consider level-$1$ and level-$2$ models in the form of \eqref{equ:levelk_policy}. Note that different human drivers may have different cognitive levels, and the autonomous ego vehicle does not know in advance the specific level, $\sigma$, of the human driver it is interacting with but has to infer $\sigma$ based on its observed information. When without any information at step $t=0$, we initialize the ego vehicle's beliefs in level-$1$ and level-$2$ models of the human-driven vehicle as $0.5$ and $0.5$.

In all of the three traffic scenarios discussed in this section, we use the following discrete-time model to represent vehicle kinematics in the longitudinal direction,
\begin{equation}
\left[\begin{matrix}
s_{t+1}^{i,j} \\
v_{t+1}^{i,j} \\
\end{matrix}\right] = \left[\begin{matrix}
1 & \Delta t  \\
0 & 1  \\
\end{matrix}\right]\left[\begin{matrix}
s_{t}^{i,j} \\
v_{t}^{i,j} \\
\end{matrix}\right] \\
+
\left[\begin{matrix}
\frac{\Delta t^2}{2} \\ \Delta t
\end{matrix}\right]
a_{t}^{i,j},
\end{equation}
where $s$ denotes position, $v$ denotes velocity, $a$ denotes acceleration, the subscript $t$ represents the discrete time, the first superscript $i \in \{1,2\}$ distinguishes the autonomous ego vehicle $1$ from the human-driven vehicle $2$, the second superscript $j \in \{x,y\}$ denotes the $x$~or~$y$-direction, and $\Delta t = 1$~[s] is the sampling period. We model lane changes as instantaneous events, i.e., completed in one time step. The acceleration $a$, taking values in a finite acceleration set $A$, and the lane change command are the actions to be decided on.

As described in Section~\ref{sec: level-k}, in order to formulate the level-$k$ models of the human-driven vehicle, $\pi^{2,k}$, we need to define the level-$0$ models of both vehicles, $\pi^{1,0}$ and $\pi^{2,0}$. Following \cite{li2018game_intersection,tian2018adaptive}, we let a level-$0$ vehicle select actions to maximize the same reward function as that for level-$1$ and $2$ vehicles but treat other vehicles on road as stationary obstacles. Note that ``as stationary obstacles'' defines a way to predict the other vehicle's actions in the decision making process \eqref{equ:DM_1}, so the ego vehicle's optimal actions, and hence the level-0 policy, can be determined.

\subsection{Intersection}

\begin{figure}[ht]
\begin{center}
\begin{picture}(200.0, 192)
\put(  0,  93){\epsfig{file=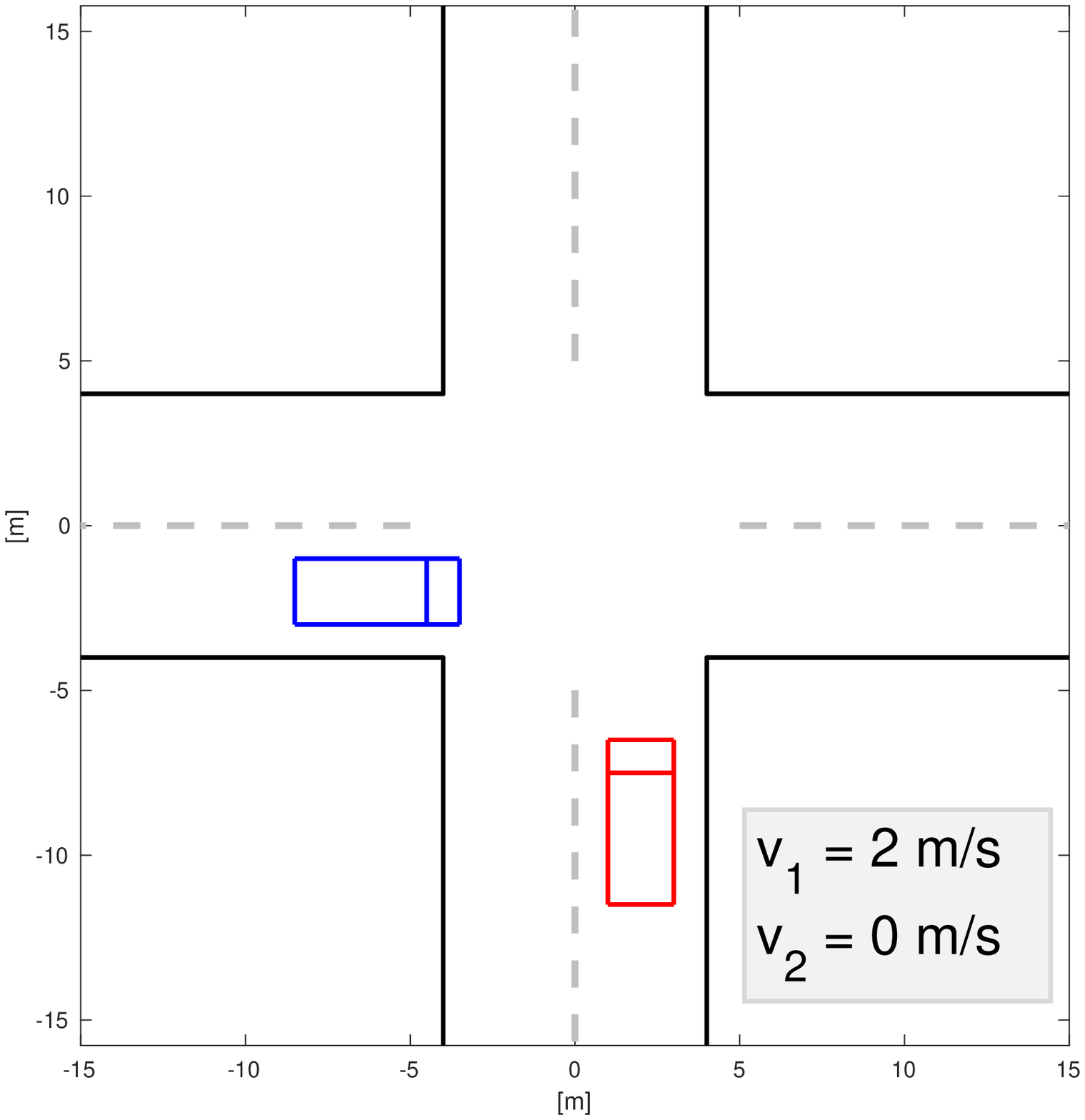,width = 0.36 \linewidth, trim=1.1cm 1cm 0cm 0cm,clip}}
\put(  110,  93){\epsfig{file=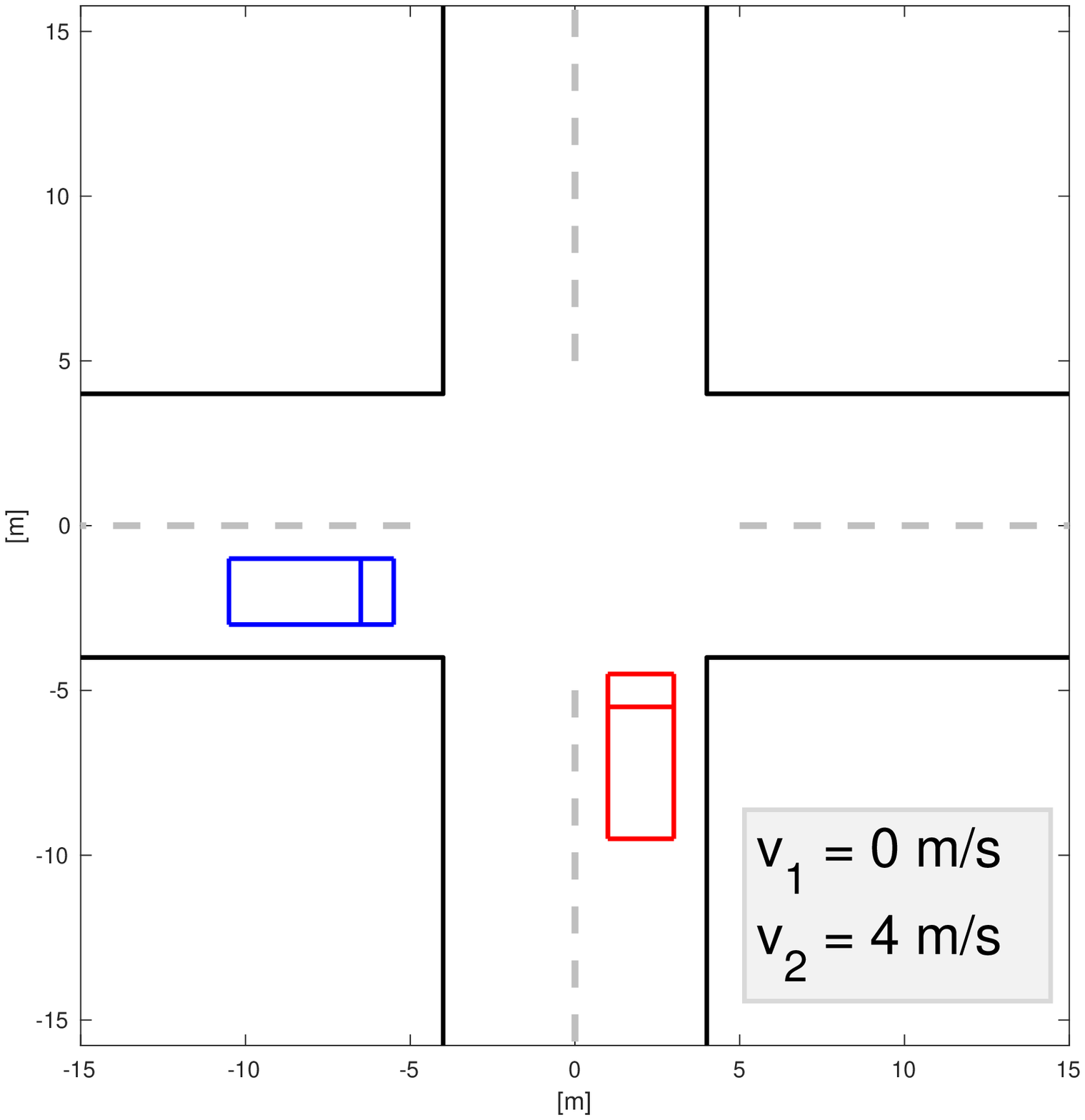,width = 0.36 \linewidth, trim=1.1cm 1cm 0cm 0cm,clip}}
\put(  0,  -2){\epsfig{file=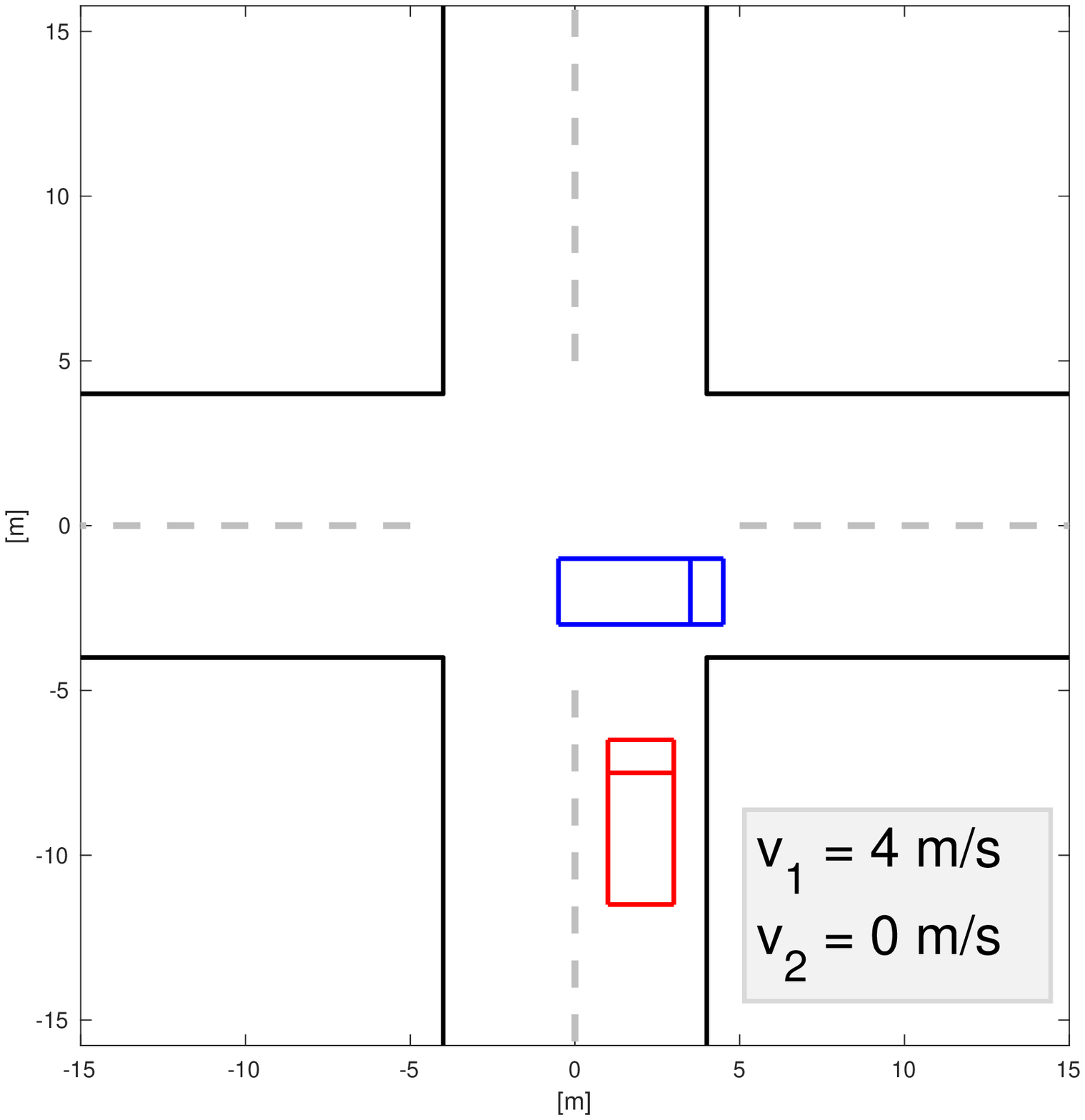,width = 0.36 \linewidth, trim=1.1cm 1cm 0cm 0cm,clip}}  
\put(  110,  -2){\epsfig{file=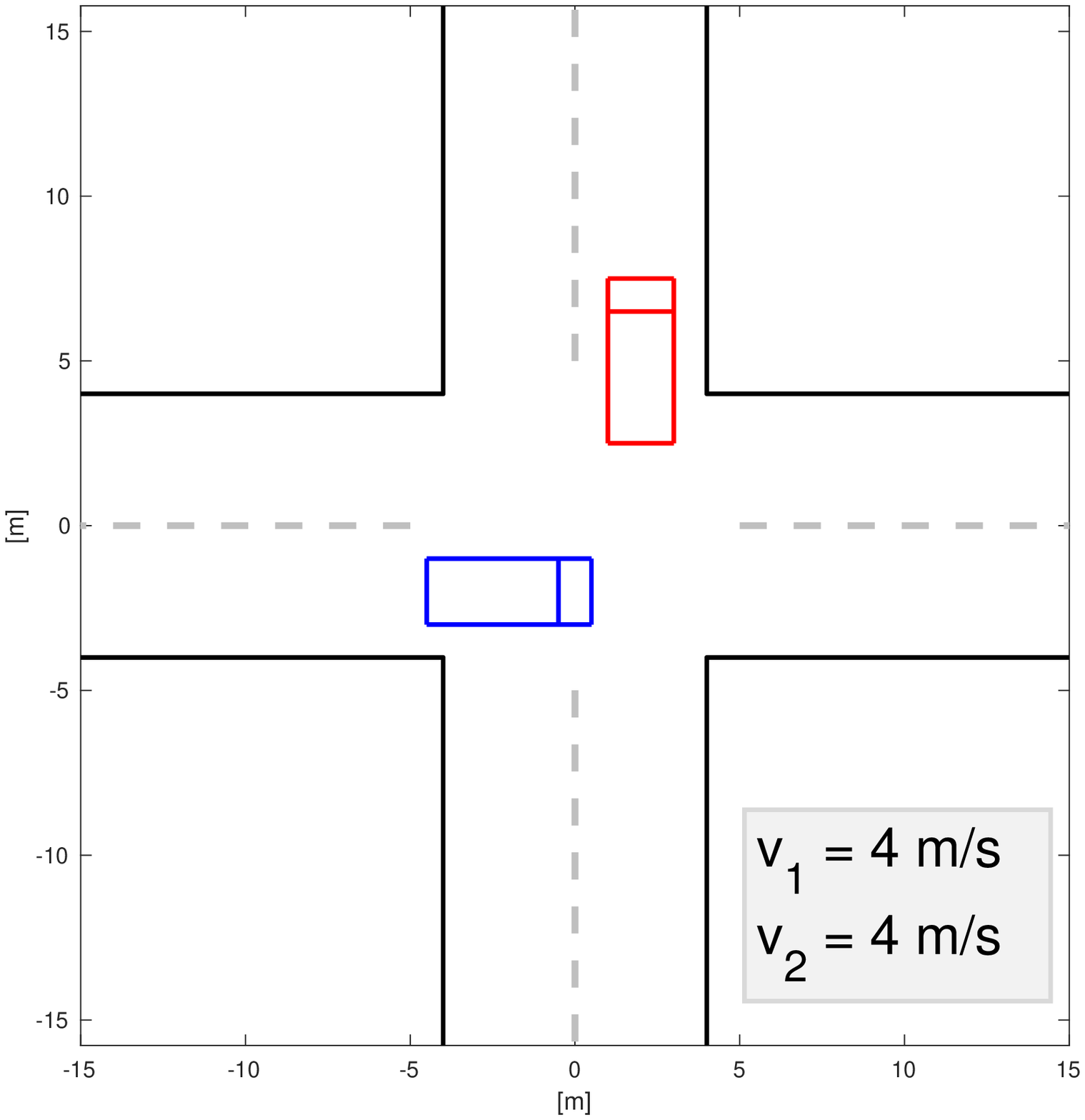,width = 0.36 \linewidth, trim=1.1cm 1cm 0cm 0cm,clip}}
\small
\put(66,173){(a-1)}
\put(176,173){(b-1)}
\put(66,78){(a-2)}
\put(176,78){(b-2)}
\normalsize
\end{picture}
\end{center}
      \caption{Intersection scenario.
      (a-1) and (a-2) show two subsequent steps in the simulation of autonomous ego vehicle (blue car) interacting with a level-$1$ human-driven vehicle (red car); (b-1) and (b-2) show those of interacting with a level-$2$ human-driven vehicle.}
      \label{fig: intersect_sim}
\end{figure}

As shown in Fig.~\ref{fig: intersect_sim}, the autonomous ego vehicle (blue car) encounters a human-driven vehicle (red car) at an unsignalized four-way intersection. Both vehicles are driving straight through the intersection. Such an objective is represented by the following reward function,
\begin{equation}
   R^i = s^{i,j},
\end{equation}
where $(i,j) = (1,x)$ for the autonomous ego vehicle and $(i,j) = (2,y)$ for the human-driven vehicle.

In the formulated receding-horizon optimization problem \eqref{equ:DM_1}, we choose the planning horizon as $N=3$, which is shown in our simulations to be effective for the autonomous ego vehicle to resolve the conflict with the human-driven vehicle in such an intersection-encounter  scenario.

Moreover, given the safety requirement to maintain the positions in the safe set,
\begin{equation}
    \Omega := \big\{ (s^1,s^2) \,|\, \| s^1 - s^2 \|_2 \geq 1.2\, l_{\text{car}} \big\},
\end{equation}
where $s^i = \big[s^{i,x},s^{i,y}\big]^\top$, $\|\cdot\|_2$ represents the Euclidean norm, and $l_{\text{car}} = 5$~[m] is the car length, we impose the following chance constraint over the planning horizon,
\begin{equation}\label{equ:AV_chance}
    \mathbb{P} \big\{(s_{\tau|t}^1,s_{\tau|t}^2) \in \Omega,\, \forall\, \tau = 1, \cdots, N \,\big|\, \xi_t \big\} \ge 0.99.
\end{equation}

Figs.~\ref{fig: intersect_sim}(a-1) and (a-2) show two subsequent steps in the simulation of autonomous ego vehicle interacting with a level-$1$ human-driven vehicle, and Figs.~\ref{fig: intersect_sim}(b-1) and (b-2) show those of interacting with a level-$2$ human-driven vehicle. When interacting with a level-$1$ human-driven vehicle, which, on the basis of our level-$0$ model introduced above, represents a cautious/conservative driver, the autonomous ego vehicle decides to drive through the intersection first. When interacting with a level-$2$ human-driven vehicle (aggressive, based on our specified level-$0$ model), the autonomous ego vehicle yields the right of way to the human-driven vehicle. The autonomous ego vehicle responds to the two different human drivers in different ways because it gains knowledge of the human driver's cognitive level by observing his/her actions for the first few steps after which it can predict his/her future actions and respond optimally.

\subsection{Overtaking}

\begin{figure}[ht]
\begin{center}
\begin{picture}(230.0, 285.0)
\put(  0,  214){\epsfig{file=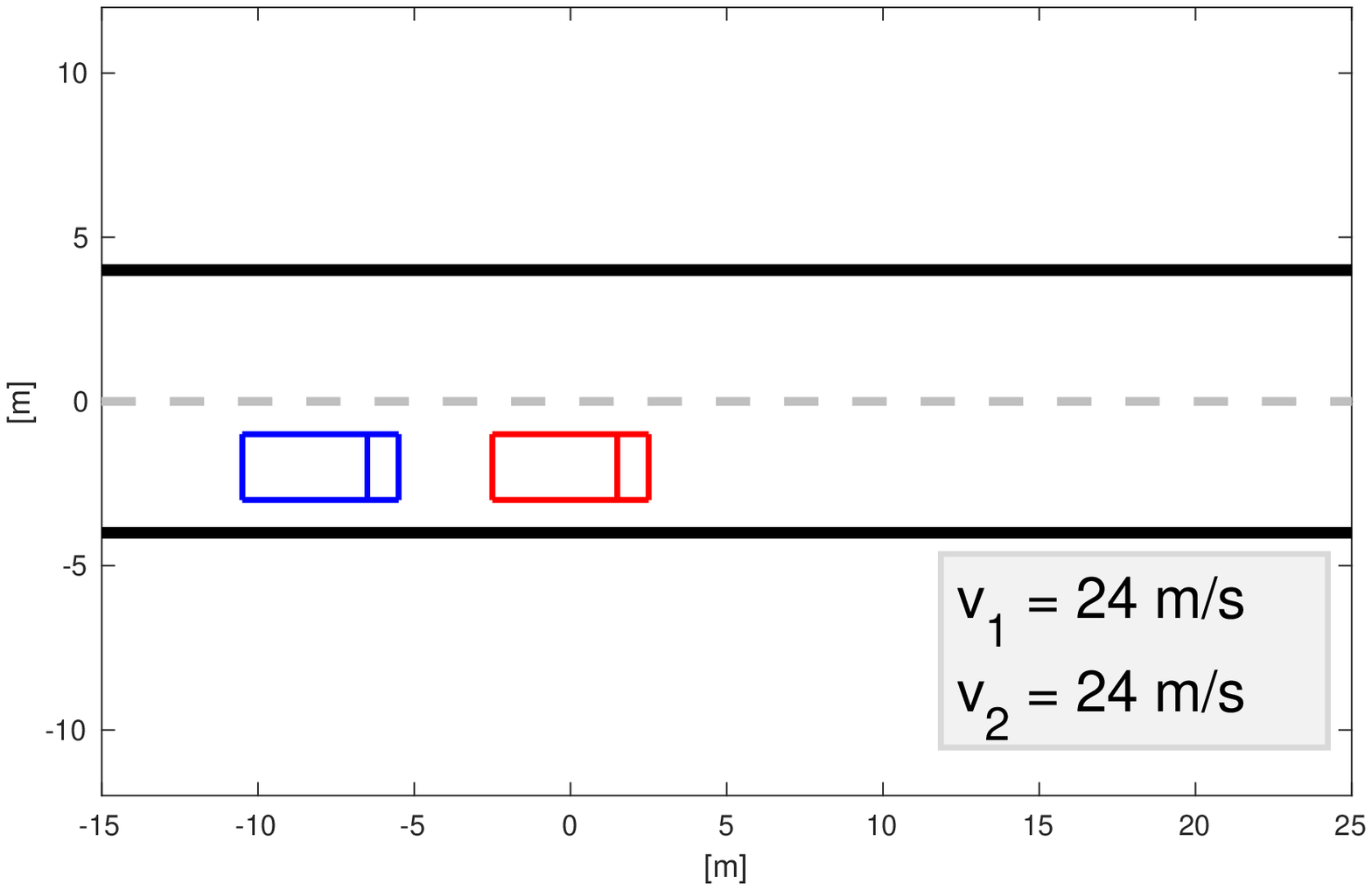,width = 0.45 \linewidth, trim=1.1cm 1cm 0cm 0cm,clip}}
\put(  120, 214){\epsfig{file=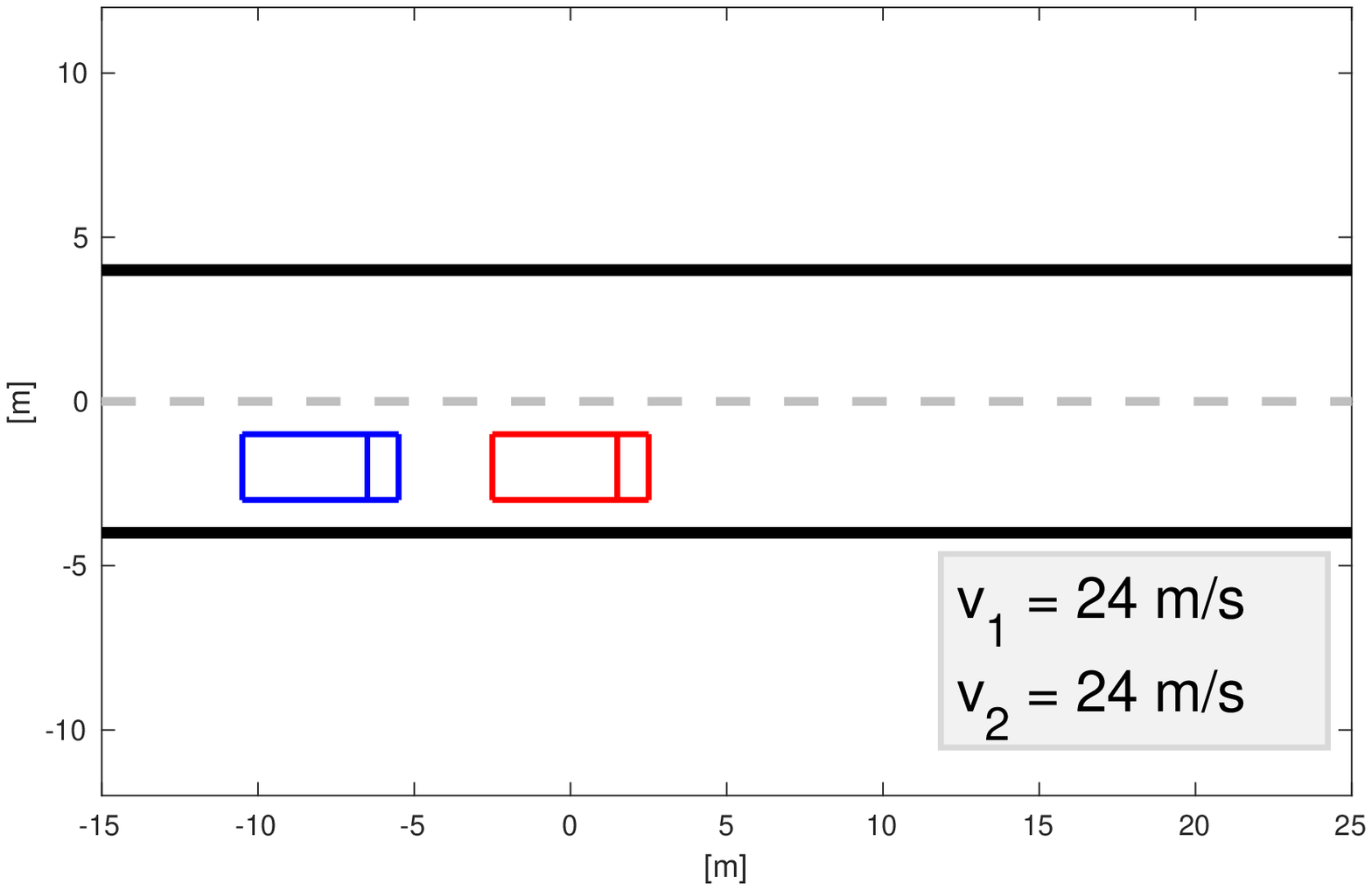,width = 0.45 \linewidth, trim=1.1cm 1cm 0cm 0cm,clip}}
\put(  0,  142){\epsfig{file=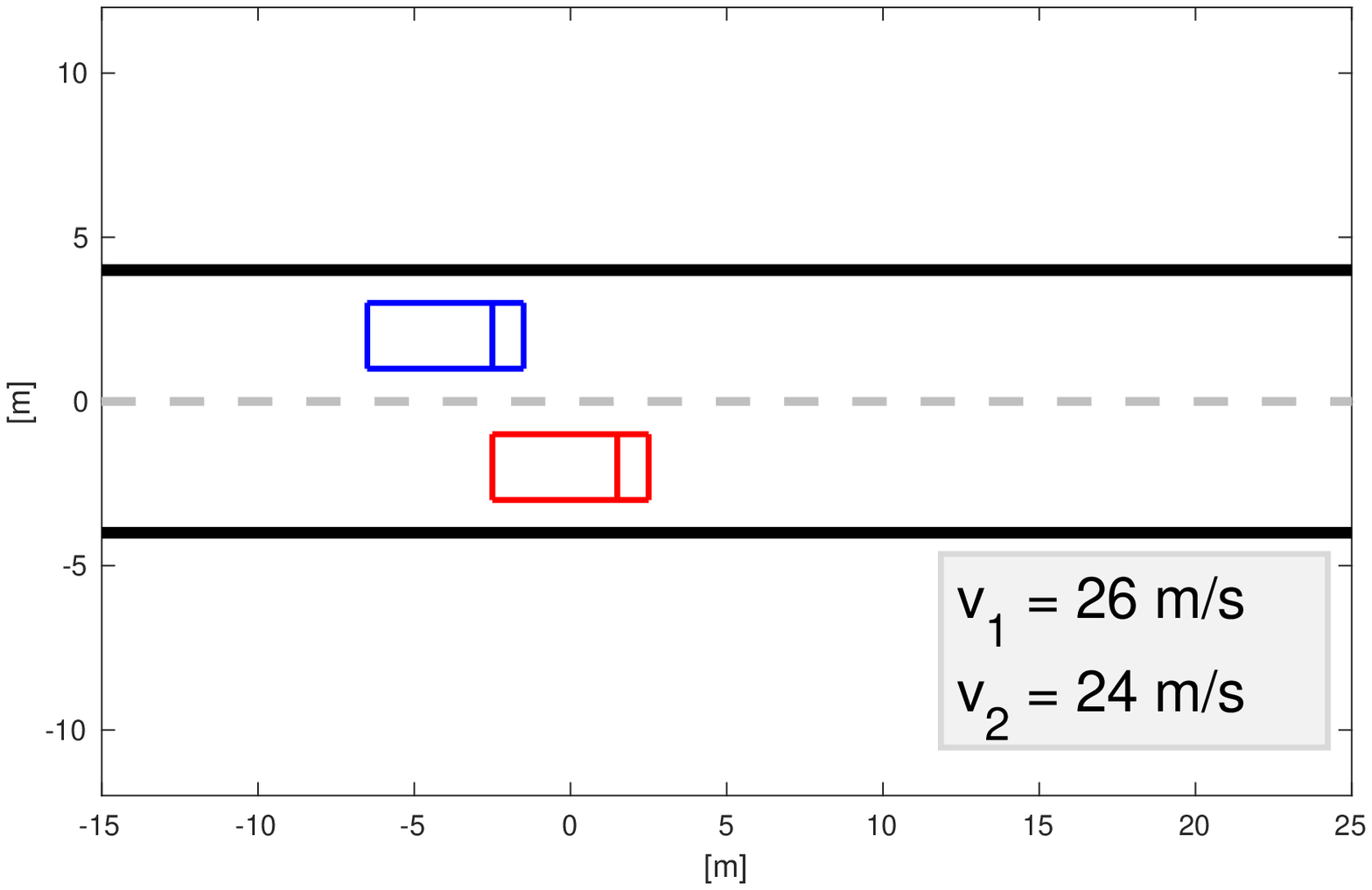,width = 0.45 \linewidth, trim=1.1cm 1cm 0cm 0cm,clip}}
\put(  120, 142){\epsfig{file=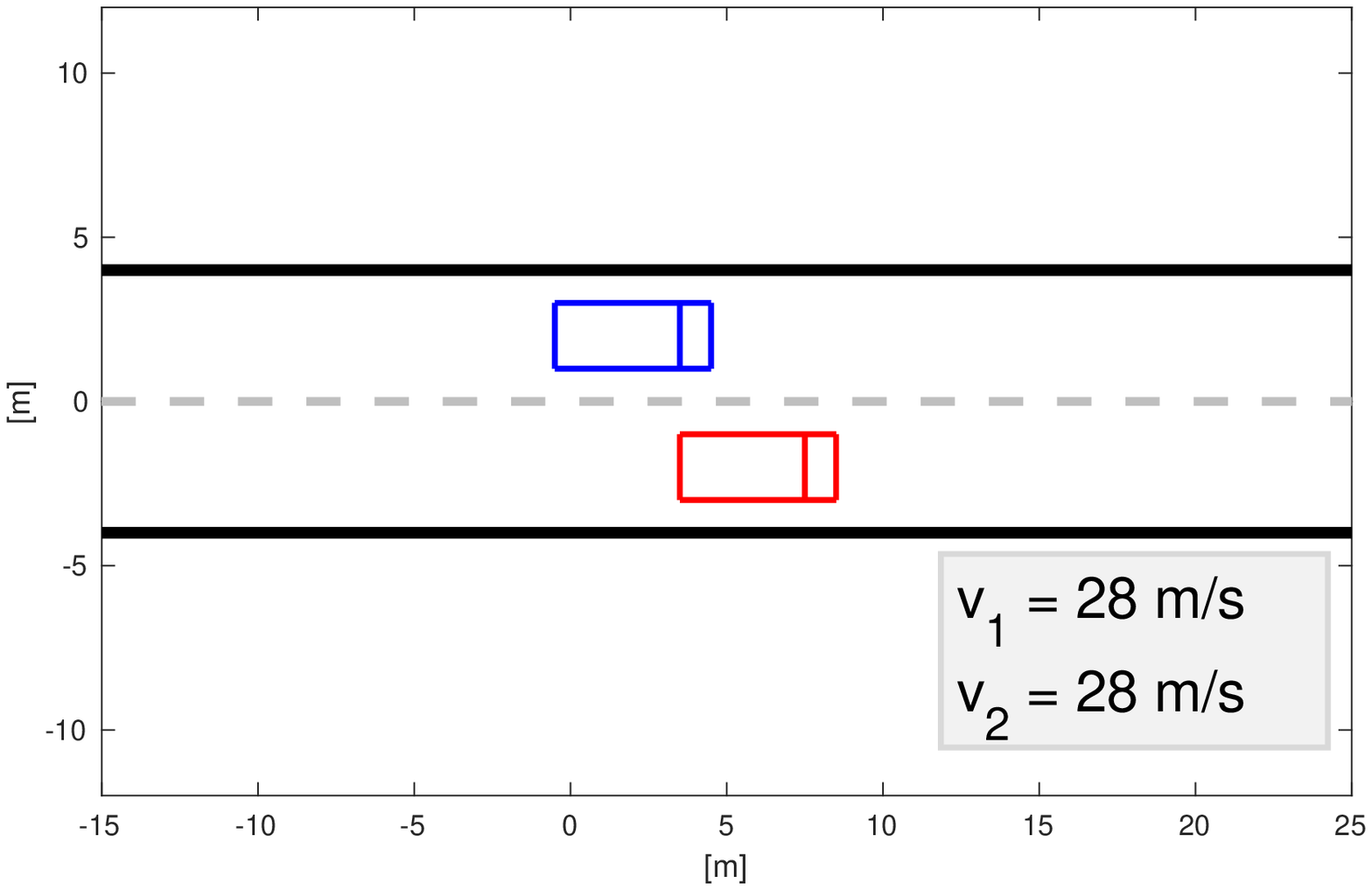,width = 0.45 \linewidth, trim=1.1cm 1cm 0cm 0cm,clip}}

\put(  0,  70){\epsfig{file=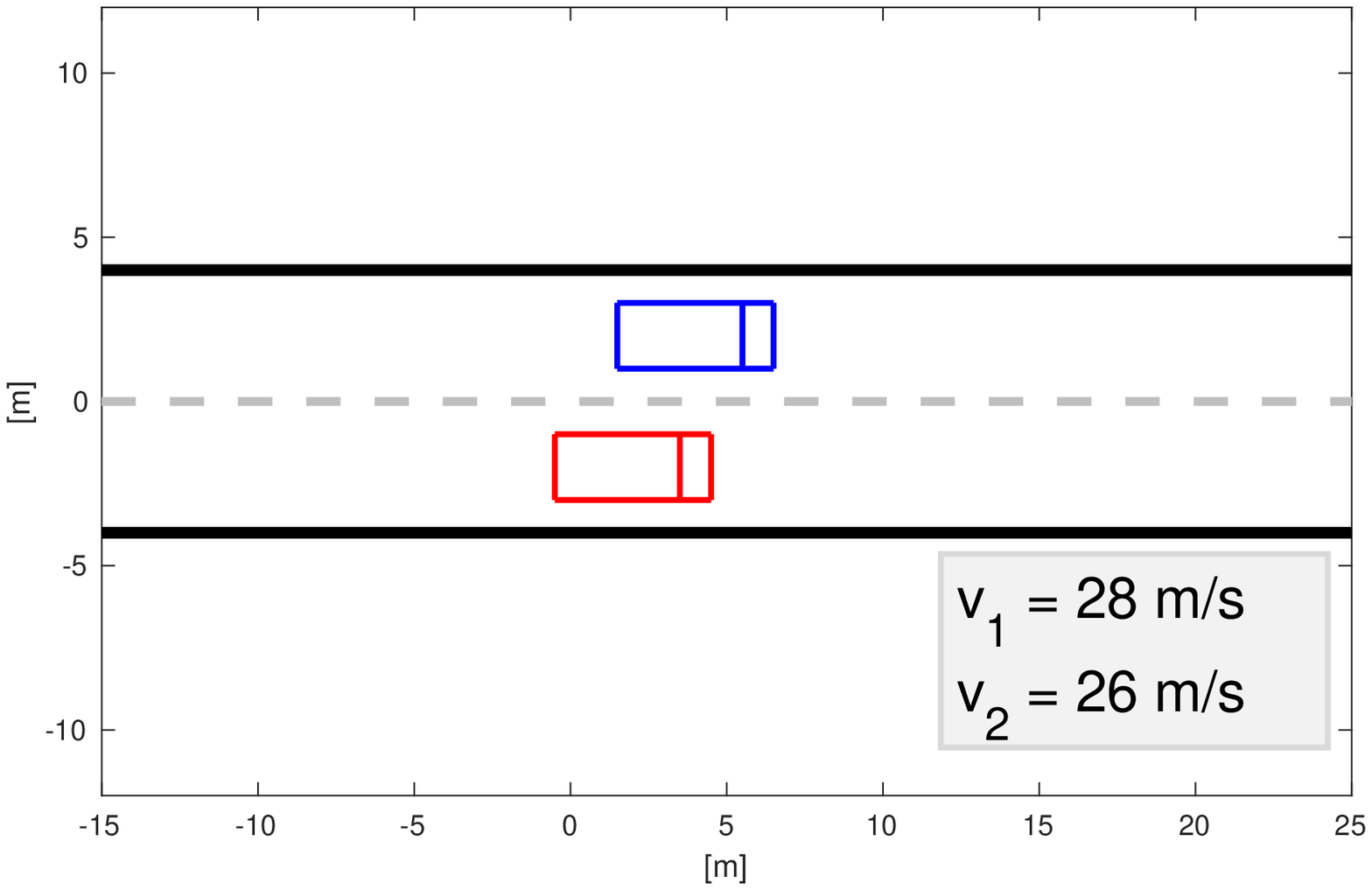,width = 0.45 \linewidth, trim=1.1cm 1cm 0cm 0cm,clip}}
\put(  120, 70){\epsfig{file=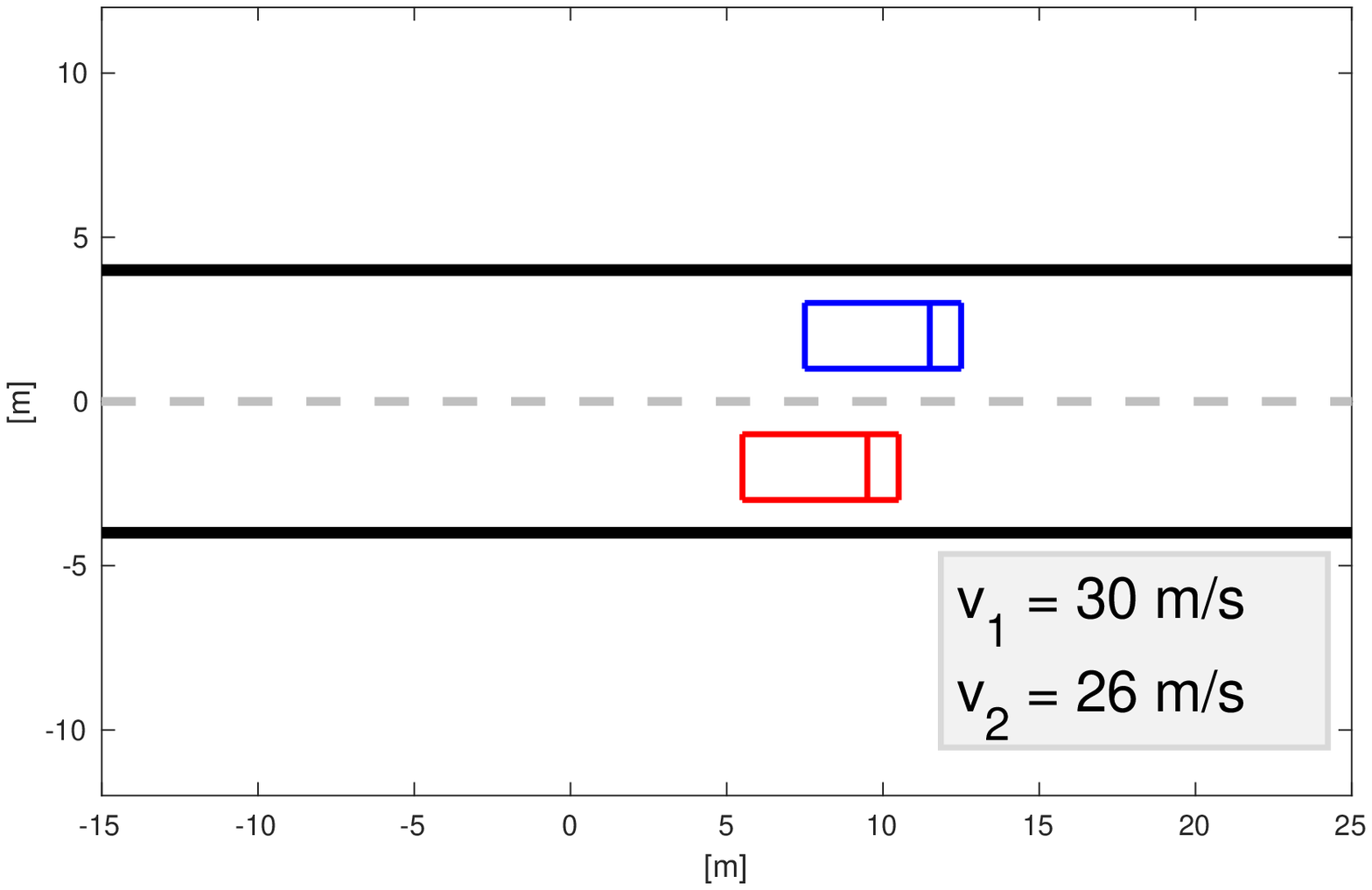,width = 0.45 \linewidth, trim=1.1cm 1cm 0cm 0cm,clip}}
\put(  0,  -2){\epsfig{file=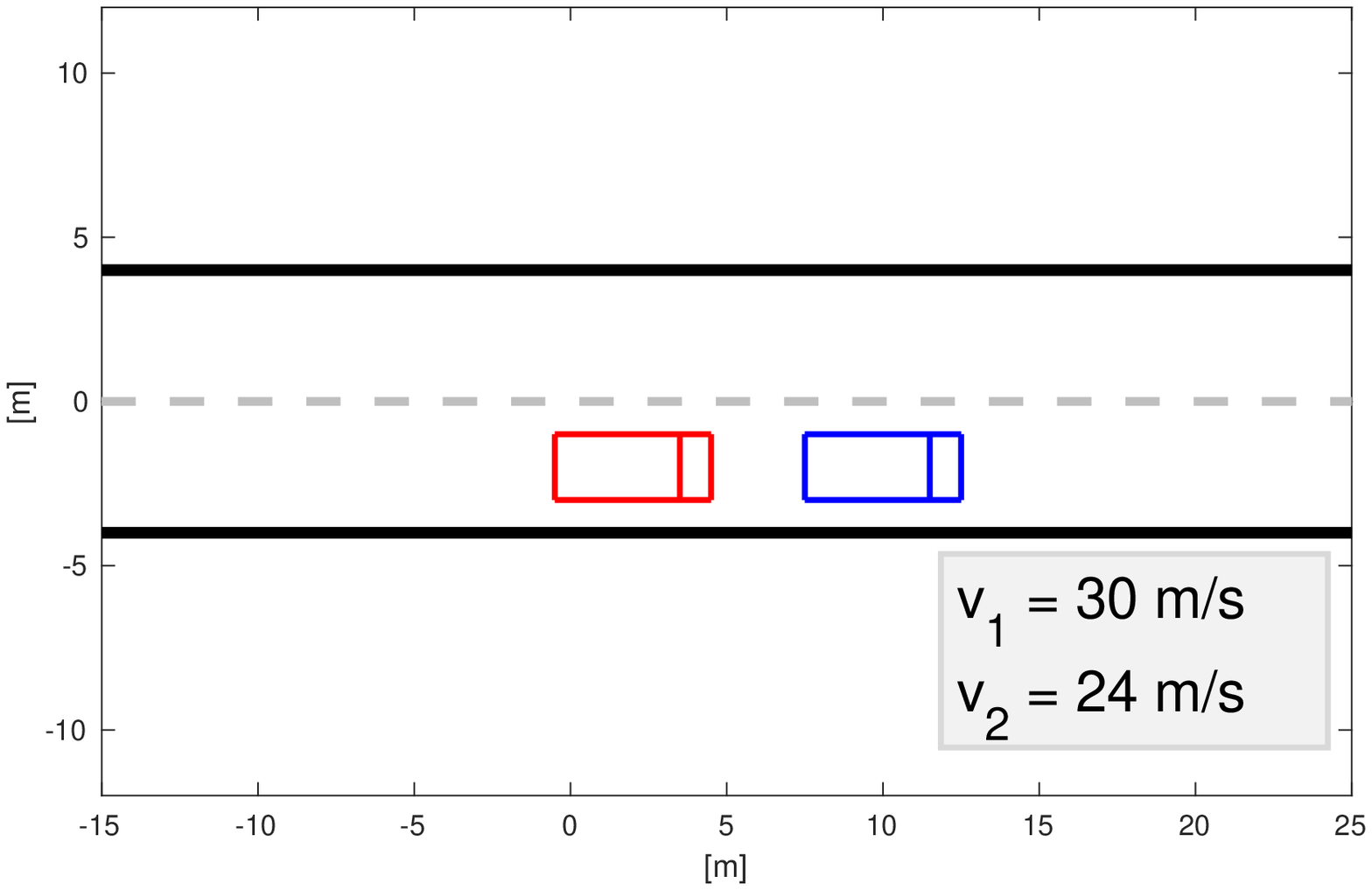,width = 0.45 \linewidth, trim=1.1cm 1cm 0cm 0cm,clip}}  
\put(  120,  -2){\epsfig{file=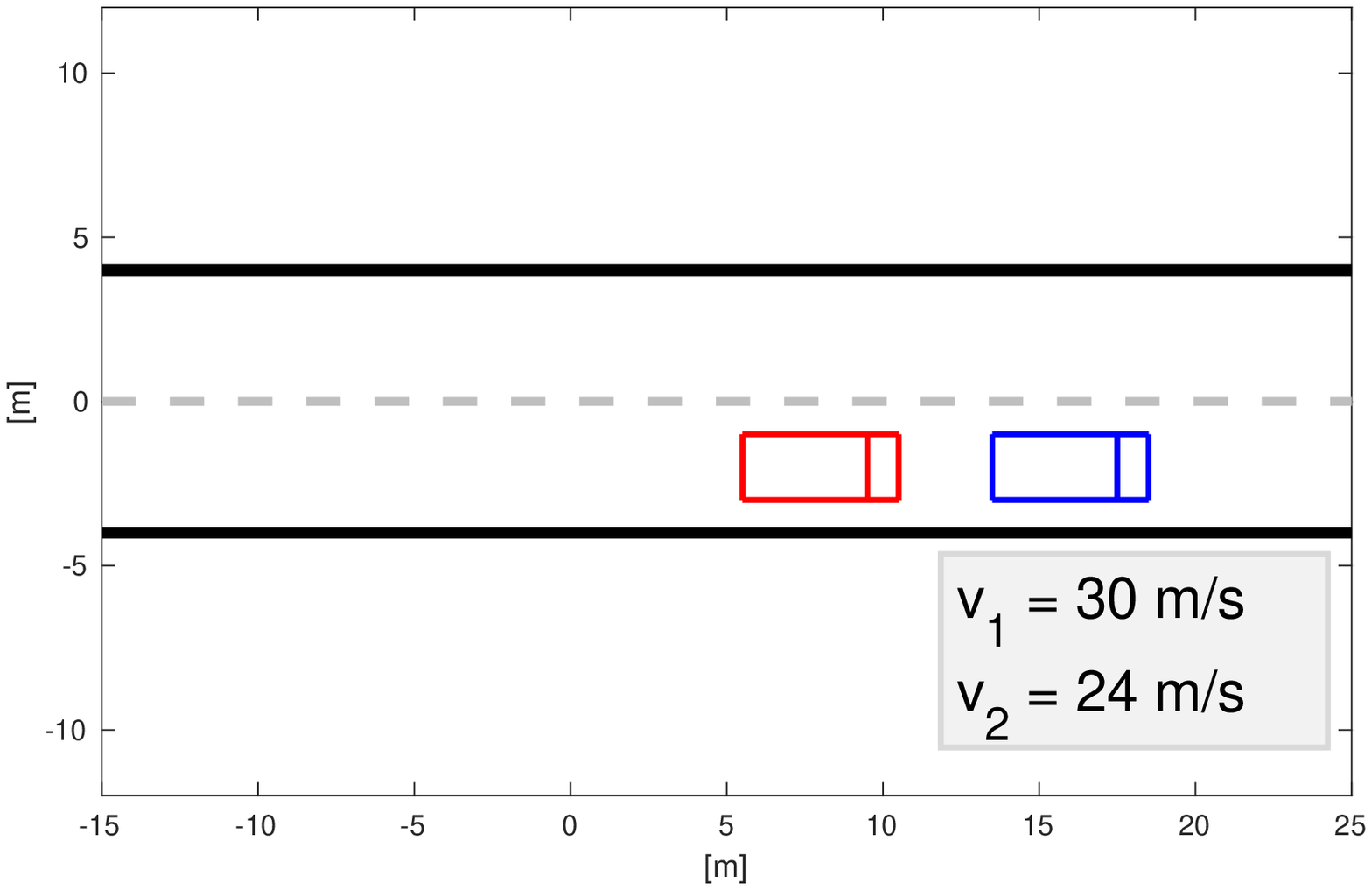,width = 0.45 \linewidth, trim=1.1cm 1cm 0cm 0cm,clip}}
\small
\put(88,273){(a-1)}
\put(208,273){(b-1)}
\put(88,201){(a-2)}
\put(208,201){(b-2)}
\put(88,129){(a-3)}
\put(208,129){(b-3)}
\put(88,57){(a-4)}
\put(208,57){(b-4)}
\normalsize
\end{picture}
\end{center}
      \caption{Overtaking scenario.
      (a-1) to (a-4) show four subsequent steps in the simulation of autonomous ego vehicle (blue car) interacting with a level-$1$ human-driven vehicle (red car); (b-1) to (b-4) show those of interacting with a level-$2$ human-driven vehicle.}
      \label{fig: overtaken_sim}
      \vspace{-0.12in}
\end{figure}

The second scenario we consider is shown in Fig.~\ref{fig: overtaken_sim}, where the autonomous ego vehicle (blue car) is overtaking a human-driven vehicle (red car). A similar scenario has been considered in \cite{li2018tractable,li2019stochastic} but not in a game theoretic setting.

We consider the following reward function,
\begin{equation}
   R^i = 8 s^{i,x} - s^{i,y},
\end{equation}
where the first term is used to encourage the autonomous ego vehicle to overtake the human-driven vehicle, and the second term is used to penalize the autonomous ego vehicle for driving in the left passing lane so that it is encouraged to come back to the right traveling lane after the overtaking as quickly as it can.

We choose the planning horizon as $N=3$ and impose a chance constraint over the planning horizon in the form of \eqref{equ:AV_chance} where the safe set $\Omega$ is now defined as
\begin{align}
    \Omega := \big\{ (s^{1},s^{2}) & \,|\, \big(| s^{1,x} - s^{2,x} | \geq 1.6\, l_{\text{car}}\big) \label{equ:safety_overtake} \\[2pt]
    &\! \vee\! \big(| s^{1,y} - s^{2,y} | \geq w_{\text{lane}}\big) \big \}, \nonumber
\end{align}
in which $w_{\text{lane}} = 3.6$~[m] is the lane width. The safe set \eqref{equ:safety_overtake} represents the requirement that overtaking can occur only when the two vehicles are traveling in different lanes, otherwise they shall keep a reasonable distance in the longitudinal direction to improve safety.

Figs.~\ref{fig: overtaken_sim}(a-1)-(a-4) show
four subsequent steps in the simulation of
autonomous ego vehicle interacting with a level-$1$ human-driven vehicle, and Figs.~\ref{fig: overtaken_sim}(b-1)-(b-4) show those of interacting with a level-$2$ human-driven vehicle. We note that in this simulation the maximum speed of the human-driven vehicle is restricted to be smaller than that of the autonomous ego vehicle to ensure the possibility of an overtaking. When interacting with a level-$1$ driver, the autonomous ego vehicle completes the overtaking relatively quickly because, as can be seen in Fig.~\ref{fig: overtaken_sim}(a-2), the level-$1$ driver drives slowly to let it cut in. When interacting with a level-$2$ driver, the autonomous ego vehicle needs to drive in the passing lane for a longer period of time before it can come back to the traveling lane.

\subsection{Merging}

\begin{figure}[h]
\begin{center}
\begin{picture}(210.0, 395)
\put(  0,
343 ){\epsfig{file=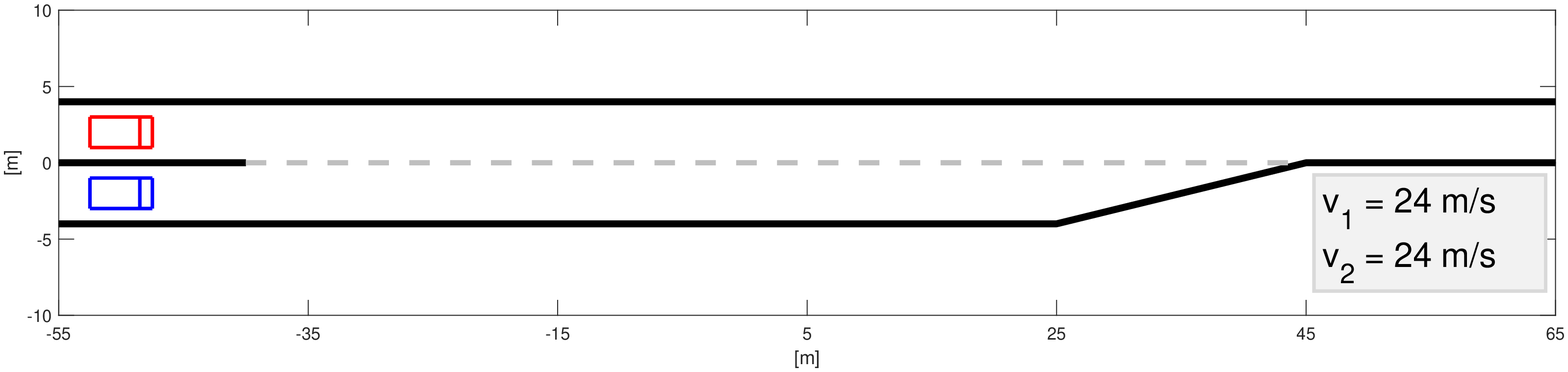,width = 0.85 \linewidth, trim=1.1cm 1cm 0cm 0cm,clip}}
\put(  0,
295 ){\epsfig{file=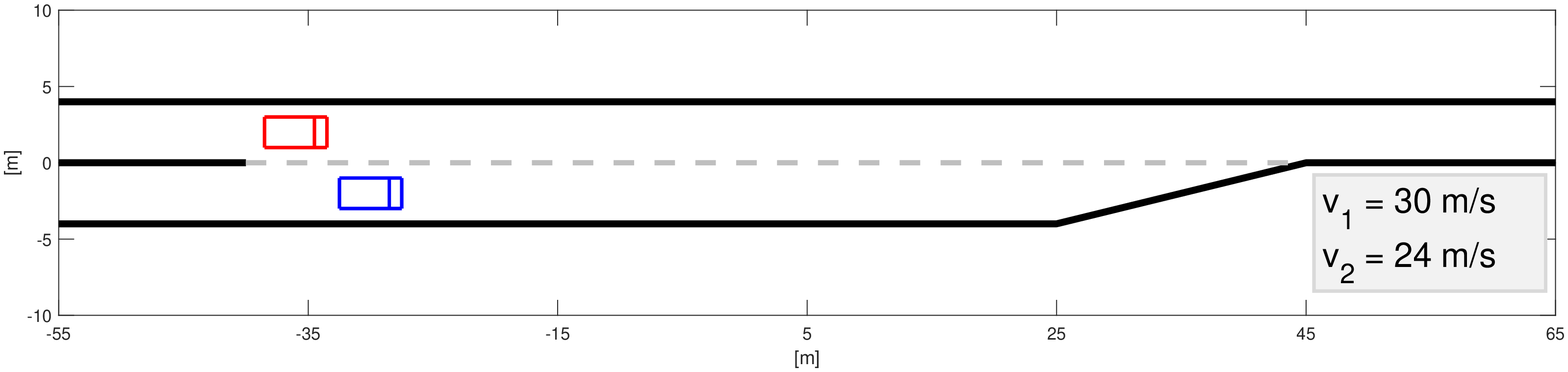,width = 0.85 \linewidth, trim=1.1cm 1cm 0cm 0cm,clip}}
\put(  0,
247 ){\epsfig{file=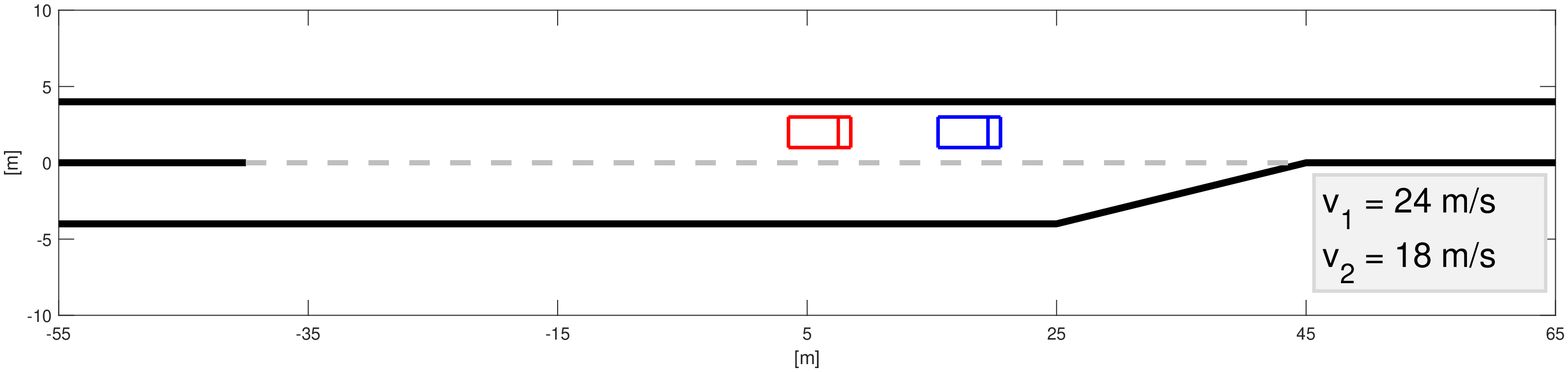,width = 0.85 \linewidth, trim=1.1cm 1cm 0cm 0cm,clip}}
\put(  0,
199 ){\epsfig{file=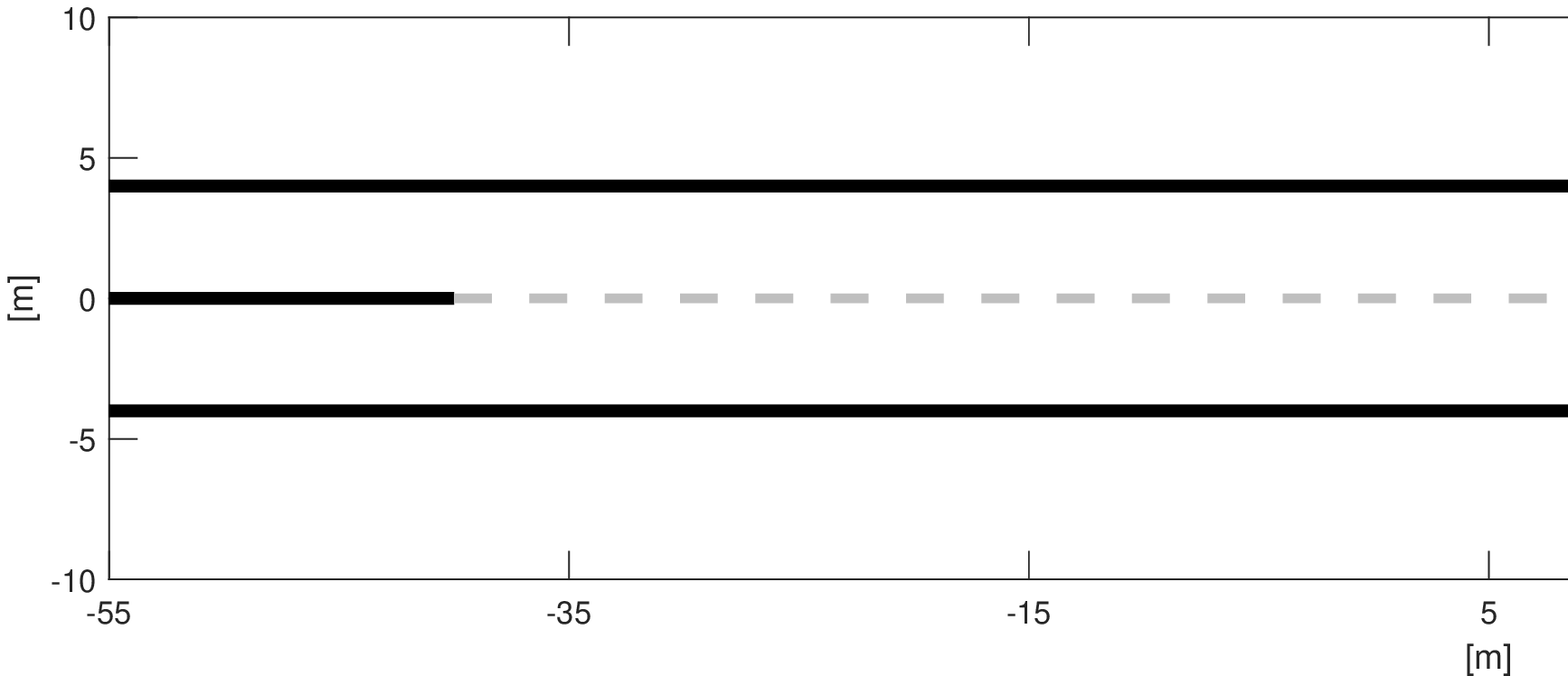,width = 0.85 \linewidth, trim=1.1cm 1cm 0cm 0cm,clip}}
\put(  0,
142 ){\epsfig{file=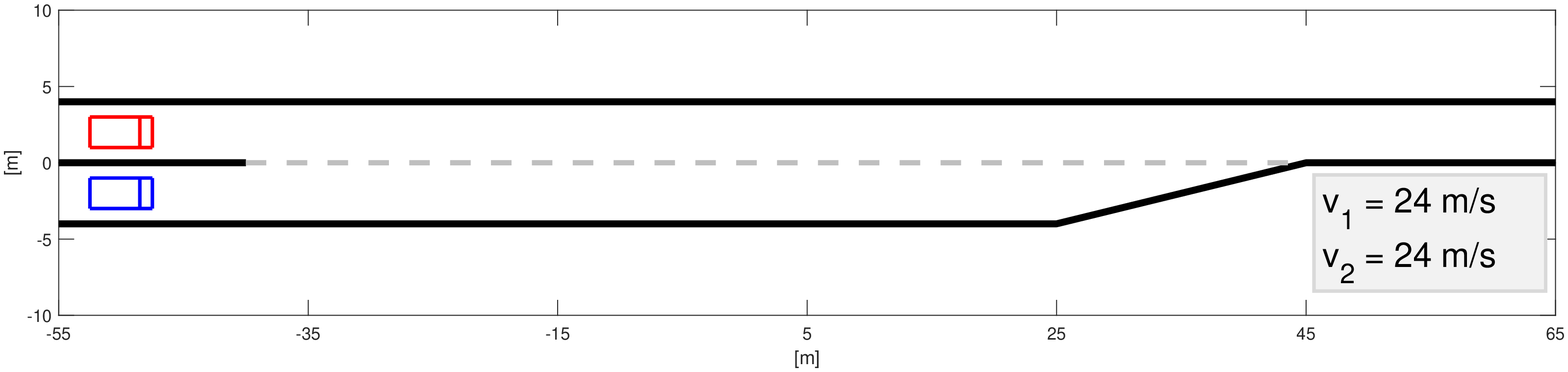,width = 0.85 \linewidth, trim=1.1cm 1cm 0cm 0cm,clip}}
\put(  0,
94 ){\epsfig{file=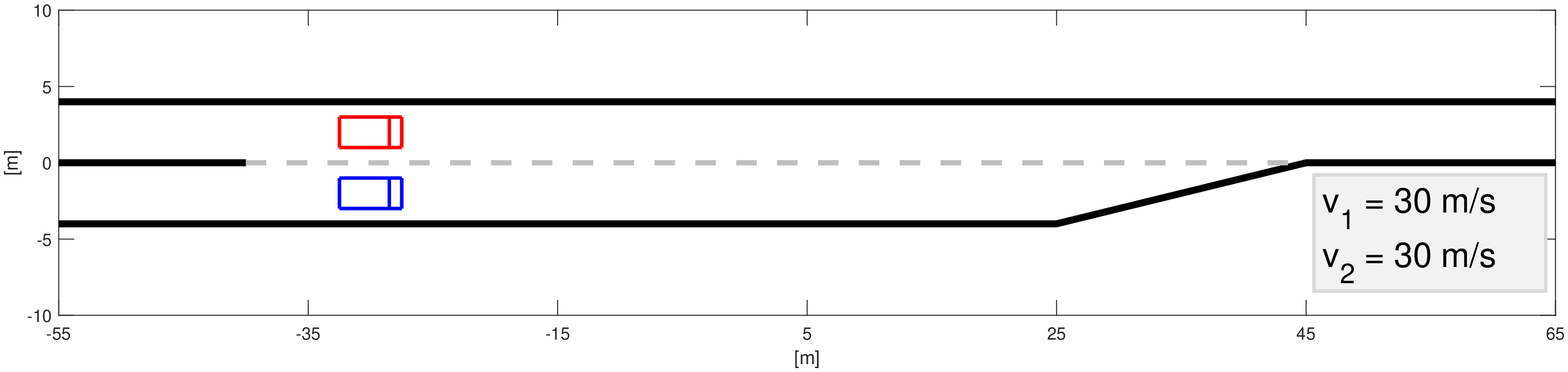,width = 0.85 \linewidth, trim=1.1cm 1cm 0cm 0cm,clip}}
\put(  0,
46 ){\epsfig{file=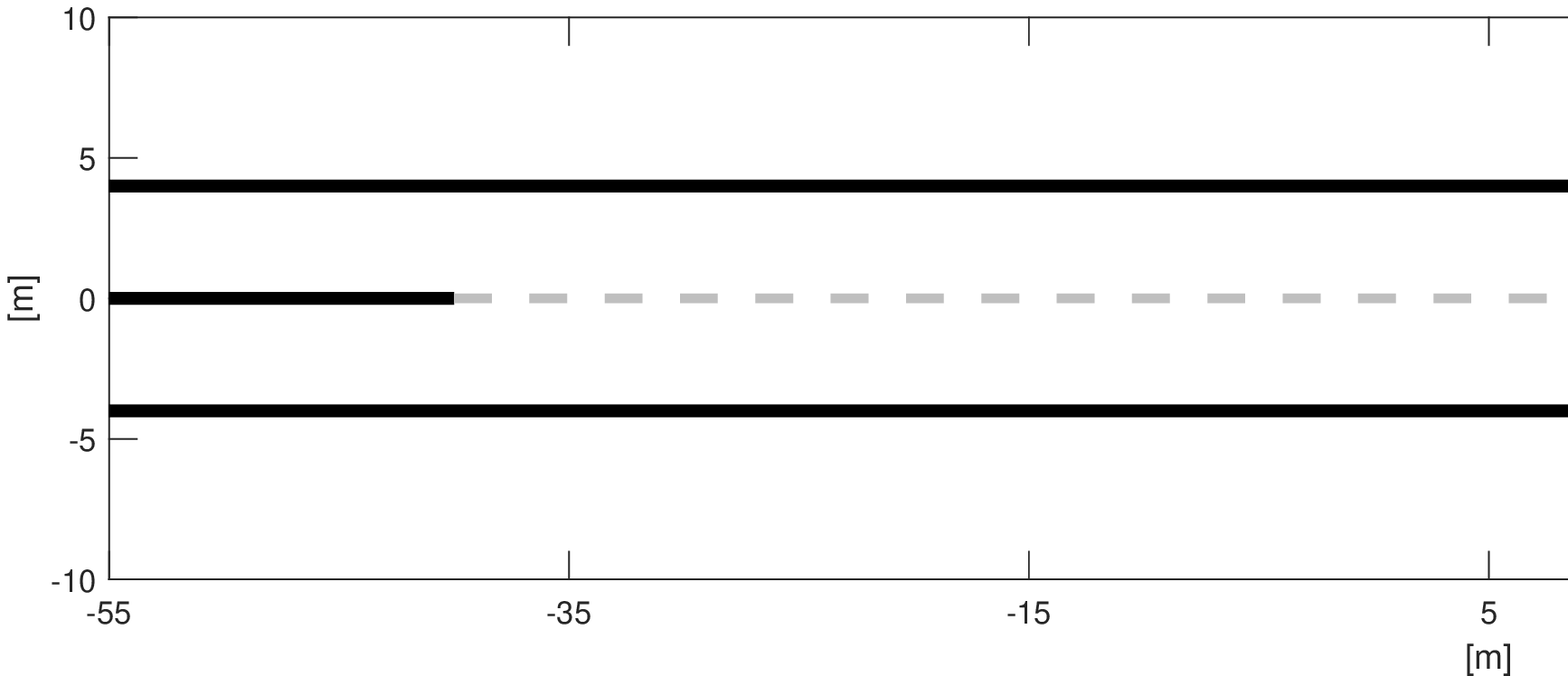,width = 0.85 \linewidth, trim=1.1cm 1cm 0cm 0cm,clip}}
\put(  0,
-2 ){\epsfig{file=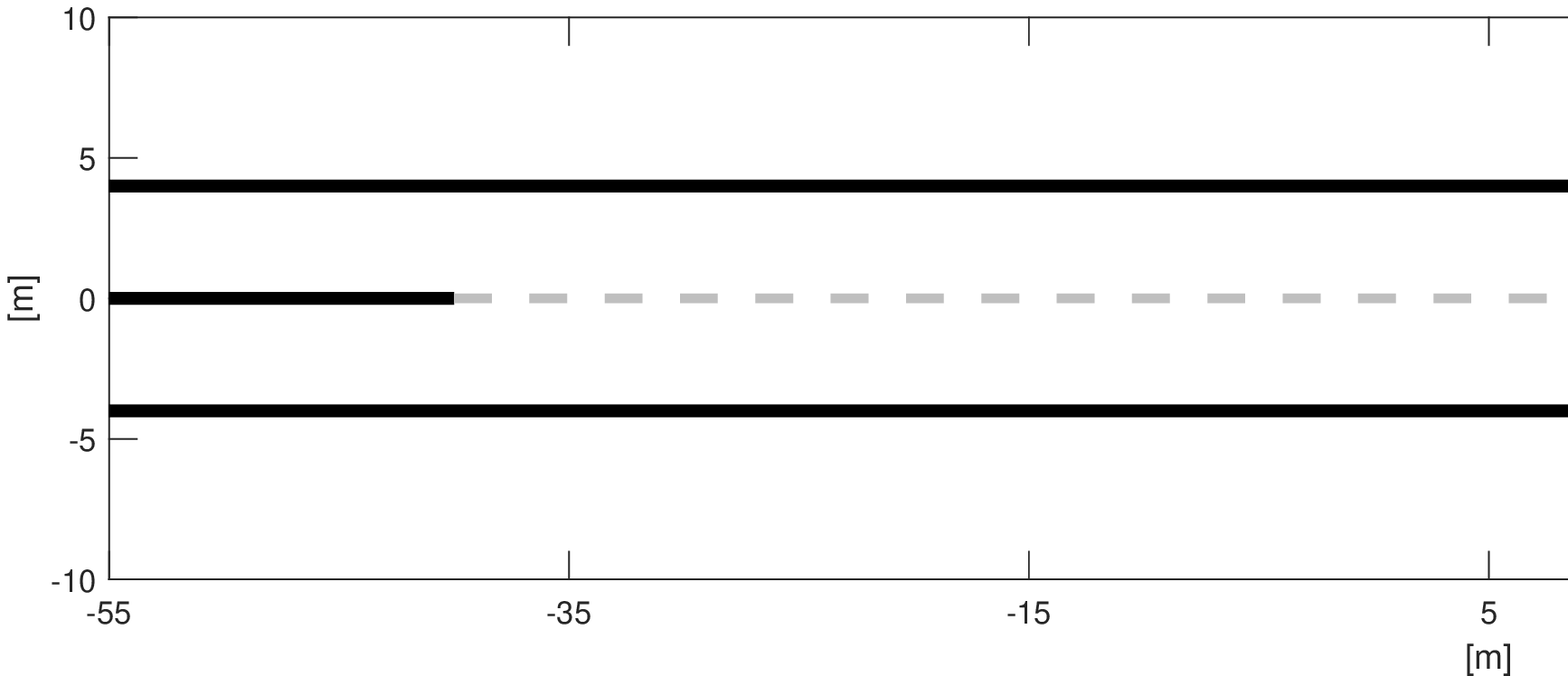,width = 0.85 \linewidth, trim=1.1cm 1cm 0cm 0cm,clip}}

\small
\put(186,377){(a-1)}
\put(186,329){(a-2)}
\put(186,281){(a-3)}
\put(186,233){(a-4)}
\put(186,176){(b-1)}
\put(186,128){(b-2)}
\put(186,80){(b-3)}
\put(186,32){(b-4)}
\normalsize
\end{picture}
\end{center}
      \caption{Merging scenario.
      (a-1) to (a-4) show four subsequent steps in the simulation of autonomous ego vehicle (blue car) interacting with a level-$1$ human-driven vehicle (red car); (b-1) to (b-4) show those of interacting with a level-$2$ human-driven vehicle.}
      \label{fig: merge_sim}
      \vspace{-0.1in}
\end{figure}

The last scenario we consider is a highway forced merging scenario. Differently from overtaking, which may improve travel speed but is usually unnecessary, merging, oftentimes, has to be accomplished within a certain road section. We consider the scenario shown in Fig.~\ref{fig: merge_sim}, where the autonomous ego vehicle (blue car) originally driving in the right lane needs to merge into the traffic in the left lane. In particular, the merging can only be and has to be accomplished within the road section with the grey-dashed lane marking.

We consider the reward function
\begin{equation}
  R^i = s^{i,x} + 10 s^{i,y}
\end{equation}
in the receding-horizon optimization \eqref{equ:DM_2}, where the first term is used to encourage the autonomous ego vehicle to maintain a reasonable travel speed and the second term is used to encourage it to merge into the left lane. For the safe set, we choose
\begin{align}
     & \Omega := \Big\{ (s^1,s^2) \,\big|\, \nonumber \\
     & \big[(| s^{1,x} - s^{2,x} | \geq 1.6\, l_{\text{car}}) \vee (| s^{1,y} - s^{2,y} | \geq w_{\text{lane}})\big] \nonumber \\[2pt]
     & \bigwedge \big[\big((s^{1,x} \leq 20) \wedge (s^{1,y} = \frac{w_{\text{lane}}}{2})\big) \nonumber \\
     &\quad\quad \vee \big(20 < s^{1,x} \leq 100\big) \label{equ:safety_merge}\\
     &\quad\quad \vee \big((s^{1,x} > 100) \wedge (s^{1,y} = \frac{3w_{\text{lane}}}{2}) \big)\big] \Big \} \nonumber
\end{align}
in the chance constraint \eqref{equ:AV_chance} so that the autonomous ego vehicle has to merge into the left lane within the road section specified by \eqref{equ:safety_merge}. The planning horizon is again chosen as $N = 3$.

Subsequent steps in the simulation with a level-$1$ human-driven vehicle (red car) traveling in the left lane are shown in Figs.~\ref{fig: merge_sim}(a-1)-(a-4), and those with a level-$2$ human-driven vehicle are in Figs.~\ref{fig: merge_sim}(b-1)-(b-4). When the human-driven vehicle is level-$1$, which, on the basis of our level-$0$ model introduced at the beginning of Section~\ref{sec:AV}, represents a cautious/conservative driver, the autonomous ego vehicle decides to merge into the left lane ahead of the human-driven vehicle. When the human-driven vehicle is level-$2$, which represents an aggressive driver, the autonomous ego vehicle merges behind the human-driven vehicle as it predicts that the human-driven vehicle will likely not yield.

\section{Discussion and conclusions}\label{sec:conclusion}

In this paper, we described a framework integrating cognitive behavioral models, Bayesian inference, and receding-horizon optimal control for autonomous decision making in a dynamic and interactive environment with uncertainty.


In the current version of the framework, the environment, which responds to the ego agent's actions, is modeled as a single intelligent agent with a certain cognitive level $\sigma$. Simulation examples representing traffic scenarios where an autonomous ego vehicle interacts with a human-driven vehicle illustrate the application of the current version of the framework. When the environment is composed of multiple intelligent agents, the proposed framework may be extended, where each of the other agents, $i = 2,3,\cdots$, is modeled separately as a level-$\sigma^i$ decision maker and the ego agent estimates each $\sigma^i$ according to agent~$i$'s historical behavior. We envision that such an extension is mainly a computational challenge rather than a theoretical one. Addressing it is left as a topic to future research.

\bibliographystyle{IEEEtran}
\bibliography{ref}

\end{document}